\documentclass[10pt,journal,compsoc]{IEEEtran}
\usepackage{graphicx} 
\usepackage{booktabs} 
\usepackage{color}
\usepackage{multirow}
\usepackage{subfigure} 
\usepackage[marginal]{footmisc}

\ifCLASSOPTIONcompsoc
\usepackage[nocompress]{cite}
\else
\usepackage{cite}
\fi
\ifCLASSINFOpdf
\else
\fi
\makeatletter
\newcommand*{\indep}{%
	\mathbin{%
		\mathpalette{\@indep}{}%
	}%
}
\newcommand*{\nindep}{%
	\mathbin{
		\mathpalette{\@indep}{\not}
	}%
}
\newcommand{\tabincell}[2]{\begin{tabular}{@{}#1@{}}#2\end{tabular}}
\newcommand*{\@indep}[2]{%
	\sbox0{$#1\perp\m@th$}
	\sbox2{$#1=$}
	\sbox4{$#1\vcenter{}$}
	\rlap{\copy0}
	\dimen@=\dimexpr\ht2-\ht4-.2pt\relax
	\kern\dimen@
	{#2}%
	\kern\dimen@
	\copy0 
}

\begin{document}
	%
	\title{Towards Privacy-Aware Causal Structure Learning in Federated Setting}
	%
	%
	%
	%

	\author{Jianli~Huang$^\dagger$,
		Xianjie~Guo$^\dagger$,
		Kui~Yu*,
		Fuyuan~Cao,
		and~Jiye~Liang
		\IEEEcompsocitemizethanks{
			\IEEEcompsocthanksitem $^\dagger$ represents equal contribution.
			\protect\\
			\IEEEcompsocthanksitem Jianli Huang, Xianjie Guo, and Kui Yu are with the Intelligent Interconnected Systems Laboratory of Anhui Province and the School of Computer Science and Information Engineering, Hefei University of Technology, Hefei 230601, China (e-mail: janelee@mail.hfut.edu.cn, xianjieguo@mail.hfut.edu.cn, yukui@hfut.edu.cn).
			
			(*Corresponding author: Kui Yu). \protect\\
			\IEEEcompsocthanksitem  Fuyuan Cao and Jiye Liang are with the Key Laboratory of Computational Intelligence and Chinese Information Processing of Ministry of Education, School of Computer and Information Technology, Shanxi University, Taiyuan 030006, China (e-mail: \{cfy, ljy\}@sxu.edu.cn). }
	}

\IEEEtitleabstractindextext{%
	\begin{abstract}
		Causal structure learning has been extensively studied and widely used in machine learning and various applications.  To achieve an ideal performance, existing causal structure learning algorithms often need to centralize a large amount of data from multiple data sources. However, in the privacy-preserving setting, it is impossible to centralize data from all sources and put them together as a single dataset. To preserve data privacy, federated learning as a new learning paradigm has attached much attention in machine learning in recent years.
		In this paper, we study a privacy-aware causal structure learning problem in the federated  setting and propose a novel Federated PC (FedPC) algorithm with two new strategies for preserving data privacy without centralizing data. Specifically, we first propose a novel layer-wise aggregation strategy for a seamless adaptation of the PC algorithm into the federated learning paradigm for federated skeleton learning, then we design an effective strategy for learning consistent separation sets for federated edge orientation. The extensive experiments validate that FedPC is effective for causal structure learning in federated learning setting. 
	\end{abstract}
	
	}

\maketitle

\IEEEdisplaynontitleabstractindextext

%
\IEEEpeerreviewmaketitle

\IEEEraisesectionheading{\section{Introduction}\label{sec:introduction}}

\IEEEPARstart{A} causal structure is often represented using a directed acyclic graph (DAG). In a DAG, a directed edge $X\rightarrow Y$ means that $X$ is a direct cause of $Y$ while $Y$ is a direct effect of $X$~\cite{pearl2009causality}. Causal structure learning aims to learn a DAG from observational data for revealing causal relations and it plays a vital role in causal inference, machine learning, and many other scientific tasks~\cite{yang2021learning,kuang2020causal,guo2022causal}. 

Various methods have been proposed for learning DAGs from observational data in the past decades~\cite{spirtes2000causation,cai2018self,nie2016learning,mooij2020joint,yang2021towards}. 
Most existing DAG learning algorithms are designed to work only on a single dataset. For achieving an ideal performance, users often need to collect data from multiple decentralized data resources and put them together as a large-scale dataset, and the large scale of data are often generated from different companies and parties. For example, a large amount of web log data of an individual user are distributed in different websites and healthcare data are owned by different hospitals.

With the increasing data privacy concerns from both governments and users, a series of data protection initiatives and laws have been proposed in recent years. Decentralized data (e.g.  healthcare and log data)  contains sensitive information and the centralizing data strategy leads to potential privacy leakage. Many data owners  prefer not to share their raw data with others owning to data privacy risk.

This makes most of DAG learning algorithms impractical in the privacy-preserving setting. 
To preserve data privacy, federated learning as a new learning paradigm has attracted  increasing attention  in machine learning~\cite{yang2019federated}. Federated learning learns a local model from each client and only sends the local model parameters to a central server for aggregating them into a global model without access to each client's data~\cite{mcmahan2017communication}.

In causal discovery area, little research has been done in developing algorithms for  privacy-aware causal structure learning with the consideration of data privacy, although there are many well-developed algorithms for causal structure learning. 

To fill this gap, in this paper, we propose to make the Peter-Clark (PC) algorithm~\cite{spirtes2000causation} fit into the federated learning setting for  privacy-aware causal structure learning.
The PC algorithm is a constraint-based and widely used algorithm for learning causal structures in various settings, and it is computationally feasible to tackle sparse DAGs with up to thousands of variables~\cite{colombo2014order}. Furthermore, most of well-established constraint-based algorithms are derived from the PC algorithm, and thus the idea of the paper can be directly applied to the PC-derived algorithms for designing privacy-aware DAG learning algorithms.

To make the PC algorithm fit into the federated learning setting, we face two challenging issues. First, since federated learning is a distributed learning paradigm while the PC algorithm is a centralized type of methods, it is challenging to seamlessly adapt the PC algorithm to the federated learning paradigm. Second, in the PC algorithm, learning separation sets is the key to orient edges. If adapting  the PC algorithm to federated learning, different clients may have different separations sets for two non-adjacent variables. It is challenging on how to identify consistent separation sets in the federated learning setting.

To tackle the issues,  our contributions are as follows.
\begin{itemize}
	\item We propose a novel federated PC (FedPC) algorithm for  privacy-aware causal structure learning in the federated learning setting. FedPC comprises two novel subroutines,  FedSkele and FedOrien, to address the above two challenging issues.
	
	\item We design the FedSkele subroutine with a novel layer-wise aggregation strategy to seamlessly adapt the PC algorithm to the federated learning paradigm for skeleton learning. This layer-wise strategy enables each client to share and update its skeleton parameters learnt at each layer of the FedPC algorithm at the server without sharing their original data. Moreover, this strategy can guarantee that the subroutine naturally converges without requiring any special parameters.
	
	\item We design the FedOrien subroutine with an effective strategy to identify consistent separation sets across clients for accurate edge orientation without centralizing data from each client to the server. 
	
	\item  We have conducted extensive experiments using synthetic, benchmark, and real datasets, and have compared FedPC with the state-of-the-art algorithms to demonstrate the effectiveness of FedPC.
\end{itemize}

\section{Related Work}\label{sec:related_work}

In the past decades, many algorithms have been designed for learning DAGs from observed data for causal structure learning~\cite{vowels2022d}. In general, existing DAG learning algorithms are categorized into three types, score-based, constraint-based, and hybrid approaches. Score-based algorithms, such as GES~\cite{chickering2002optimal} and GGSL~\cite{gao2017local}, use a scoring function and a greedy search method to learn a DAG with the highest score by searching over all possible DAGs in a dataset and the representative algorithms.
Constraint-based methods, such as PC and FCI~\cite{spirtes2000causation}, employ conditional independence (CI) tests to first learn a skeleton of a DAG, then orient the edges in the skeleton. The hybrid methods, such as MMHC~\cite{tsamardinos2006max}, BCSL~\cite{guo2022bootstrap} and ADL~\cite{guo2023adaptive}, are the combinations of score-based and constraint-based methods. 
The score-based and constraint-based methods often return a completed partially DAG (CPDAG), i.e., a Markov equivalence class.

Zheng et al.~\cite{zheng2018dags} recently proposed to learn DAGs using gradient-based methods and designed the NOTEARS algorithm. NOTEARS defines a DAG as a weighted adjacency matrix and formulates the acyclic constraint as an equality acyclicity constraint and find the DAG using gradient-based methods with least squares loss. NOTEARS is designed under the assumption of the linear relations between variables. \
Subsequent works have extended NOTEARS to handle nonlinear cases, such as GOLEM\cite{ng2020role} and DAG-NoCurl\cite{yu2021dags}. Another kind of methods employ different types of deep neural networks and a series of algorithms for learning DAGs. For example, DAG-GNN\cite{yu2019dag}, GraN-DAG\cite{lachapelle2020gradient}, NOTEARS-MLP\cite{zheng2020learning} are some of the methods proposed. 
Moreover, the MCSL (Masked gradient-based Causal Structure Learning) algorithm~\cite{ng2022masked} is also a gradient-based causal structure learning method. By using the binary mask, the results are mostly near either zero or one, so that the edges are easily identified by the threshold.
We refer readers to the recent two survey papers of DAG learning for more details~\cite{glymour2019review,vowels2022d}.

In addition to the algorithms designed to work only on a single dataset, there is a line of work that is able to learn causal structures directly from multiple datasets. On the one hand, most of these methods require the multiple datasets with non-identical sets but sharing a small common set of variables~\cite{huang2020causal,triantafillou2015constraint}. On the other hand, all of these algorithms do not consider data privacy problems.

Recently, Xiong et al.~\cite{xiong2021federated} presented a causal inference algorithm in the federated setting, but this algorithm does not learn causal structures. It aims to calculate causal effects. 
The closely related federated DAG learning algorithms using gradient-based optimization are FedDAG proposed by Gao et al.\cite{gao2023feddag} and NOTEARS-ADMM proposed by Ng and Zhang~\cite{ng2022towards}.
NOTEARS-ADMM uses the alternating direction method of multipliers (ADMM), such that only the model parameters have to be exchanged during the optimization process and is capable of handling both linear and nonlinear cases. In contrast, FedDAG proposes a two-level structure consisting of a graph structure learning part and a mechanism approximating part, separately learns the mechanisms on local data and jointly learns the DAG structure to handle the data heterogeneity elegantly. Furthermore, the FedDAG method is based on MCSL [21] that uses binary masks, resulting in a binary adjacency matrix instead of a real-valued one. However, these methods often require employing complex neural network models or optimization techniques to achieve satisfactory performance, which can result in high computational overhead. In addition, although existing methods perform well on synthetic linear or nonlinear datasets, they may not be effective on discrete datasets.

In summary, many algorithms have been proposed for learning DAGs, but few methods have been designed for learning causal structures by considering data privacy problems. In this paper, we propose to develop new algorithms of causal structure learning by considering data privacy.

\section{Methodology}\label{sec:methodology}
In this section, we first briefly describe the original PC algorithm in Section~\ref{sec3-1}, then present our proposed FedPC algorithm in Section~\ref{sec3-2}, and finally discuss the privacy and costs of FedPC in Section~\ref{sec3-3}.
Table~\ref{Summary-Notations} provides a summary of the notations frequently used in this paper.


\begin{table}
	\caption{Summary of Notations.}
	\centering
	\scriptsize
	\begin{tabular}{lll}
		\toprule
		& Notation & Meaning \\
		\midrule
		& ${\cal X}$      & the set of variables in a dataset      \\
		& $X_i,X_j$      & a single variable in $\mathcal{X}(i,j=1,2,...,m)$        \\
		& $N$	& number of clients\\
		& $X_i\indep X_j|\textit{\textbf{Z}}$	& $X_i$and $X_j$ are conditionally dependent given $\textit{\textbf{Z}}$\\
		& $X_i\nindep X_j|\textit{\textbf{Z}}$	& $X_i$and $X_j$ are conditionally independent given $\textit{\textbf{Z}}$\\
		& $\textit{\textbf{ne}}(\mathcal{S},X_i)$	& the set of direct neighbors of $X_i$ in $\mathcal{S}$\\
		& $\textit{\textbf{SepSet}}(X_i,X_j)$	& a separation set of $X_i$ from  $X_j$ \\
		& $\ell$	& the size of a separation set\\
		& $\textbf{\textit{Z}}$	& a separation set with the length $\ell$\\ 
		& $\mathcal{S}$	& a direct acyclic graph over ${\cal X}$\\
		& $\mathcal{S}^c$	& the complete undirected graph over $\mathcal{X}$ \\
		& $\mathcal{S}^{\ell}$	& the skeleton $\mathcal{S}^{\ell}$ obtained at the $\ell$-th layer\\ 
		& $\mathcal{S}^*$	& the final skeleton obtained in FedSkele subroutine\\
		& $<X_i,X_k,X_j>$	& an unshield triple $X_i-X_k-X_j$ in the skeleton \\		
		& $\alpha $ & the significance level of the statistical test\\
		\bottomrule
	\end{tabular}\label{Summary-Notations}
\end{table}

\subsection{The PC algorithm}\label{sec3-1}

Let $\mathcal{X}=\{X_1,X_2,...,X_m\}$ be the set of random variables in a dataset, $\mathcal{S}$  an undirected graph representing the skeleton of a DAG over $\mathcal{X}$, and $\textit{\textbf{ne}}(\mathcal{S},X_i)$  the set of direct neighbors of $X_i$ in $\mathcal{S}$, and $\mathcal{S}^c$  the complete (fully connected) undirected graph over $\mathcal{X}$. $\textit{\textbf{SepSet}}(X_i,X_j)$ is a separation set (conditioning set) that makes $X_i$ and $X_j$ conditionally independent, and $X_i\indep X_j|\textit{\textbf{Z}}$ ($X_i\nindep X_j|\textit{\textbf{Z}}$) represents that given a conditioning set \textbf{\textit{Z}}, $X_i$ and $X_j$ are conditionally independent (dependent). The size of a conditioning set is denoted as $\ell$. An unshielded triple $<X_i,X_k,X_j>$ in $\mathcal{S}$ denotes the local skeleton $X_i-X_k-X_j$ where $X_i$ and $X_j$ are not directly adjacent, but $X_k$ is a direct neighbour of $X_i$ and $X_j$. An unshielded triple $<X_i,X_k,X_j>$ forms a v-structure, i.e.,  $X_i\rightarrow X_k\leftarrow X_j$, if both $X_i\indep X_j|\textbf{\textit{Z}}$ and $X_k\notin \textbf{\textit{Z}}$ hold.

The well-known PC algorithm~\cite{spirtes2000causation} as a constraint-based method consists of the following three steps.
Step 1 takes $\mathcal{S}^c$ as the initial skeleton, then updates the skeleton layer by layer by conducting conditional independence (CI) tests with an increasing value of $\ell$. 

The value of $\ell$ initially is set 0. 
At the first layer where $\ell=0$ and $\mathcal{S}^c$ is taken as the initial skeleton, all pairs of adjacent variables in $\mathcal{S}^c$ ($\mathcal{S}^c$ is iteratively updated at this layer)  are tested with an empty conditioning set $\textbf{\textit{Z}}$ (i.e. $\ell=|\textbf{\textit{Z}}|=0$). If $X_i\indep X_j$ holds, the edge between $X_i$ and $X_j$ is removed from $\mathcal{S}^c$ and the empty set $\textit{\textbf{Z}}$ is saved as a separation set in both $\textit{\textbf{SepSet}}(X_i,X_j)$ and $\textit{\textbf{SepSet}}(X_j,X_i)$. After all pairs of adjacent variables have been checked, the algorithm proceeds to the next layer with $\ell=1$, and the skeleton obtained at the end of this layer is marked as $\mathcal{S}^0$.
Then at the layer with $\ell=1$ and $\mathcal{S}^0$ as the initial skeleton, the algorithm chooses a pair of adjacent variables $(X_i,X_j)$ in $\mathcal{S}^0$, and checks whether there exsits a subset $\textit{\textbf{Z}}\subseteq \textbf{\textit{ne}}(\mathcal{S}^0,X_i)\setminus\{X_j\}$ with $|\textbf{\textit{Z}}|=1$ which makes $X_i\indep X_j|\textit{\textbf{Z}}$ hold. If so, the edge between $X_i$ and $X_j$ is deleted from  $\mathcal{S}^0$  and the conditioning set $\textit{\textbf{Z}}$ is saved in the separation sets for both $X_i$ and $X_j$. If all pairs of adjacent variables have been checked, the algorithm increases $\ell$ by one up to 2 and the skeleton obtained at the layer is marked as $\mathcal{S}^1$. This process continuous layer by layer until the sizes of direct neighbors of all variables in the skeleton of current layer are smaller than $\ell$. We record the final skeleton as $\mathcal{S}^*$.

Step 2 identifies the v-structures by considering all unshielded triples in $\mathcal{S}^*$, and orients an unshielded triple $<X_i,X_k,X_j>$ as a v-structure if and only if $X_k\notin\textit{\textbf{SepSet}}(X_i,X_j)$. Finally, Step 3 orients as many of the remaining undirected edges as possible using the Meek rules ~\cite{meek1995causal}. The Meek rules require that orienting a remaining undirected edge does not form a new v-structure or a directed cycle in the current structure.

\subsection{The proposed FedPC algorithm}\label{sec3-2}

Since the PC algorithm works only on a single dataset, a simple strategy to adapt the PC algorithm into federated learning is to learn a skeleton at each client independently, then aggregate the learnt skeletons and orient edges at a central server. 
However, the strategy has two potential problems. First, the different qualities of data (e.g. noise or small-sized samples) at different clients may lead to learnt skeletons with highly varying qualities, then directly aggregating them  may not get a satisfactory final skeleton.
Second, during skeleton learning, when $X_i$ and $X_j$ are conditionally independent, they may have a different separation set at each client due to the data quality problems, while an inaccurate separation set will lead to errors in edge orientations.

To protect data privacy and tackle the above two problems, we have designed the FedPC algorithm to work in the federated setting and have two subroutines, FedSkele (Federated Skeleton construction) and FedOrien (Federated edge Orientation).	
\begin{figure}[htbp]
	\centering
	\includegraphics[width=0.42\textwidth]{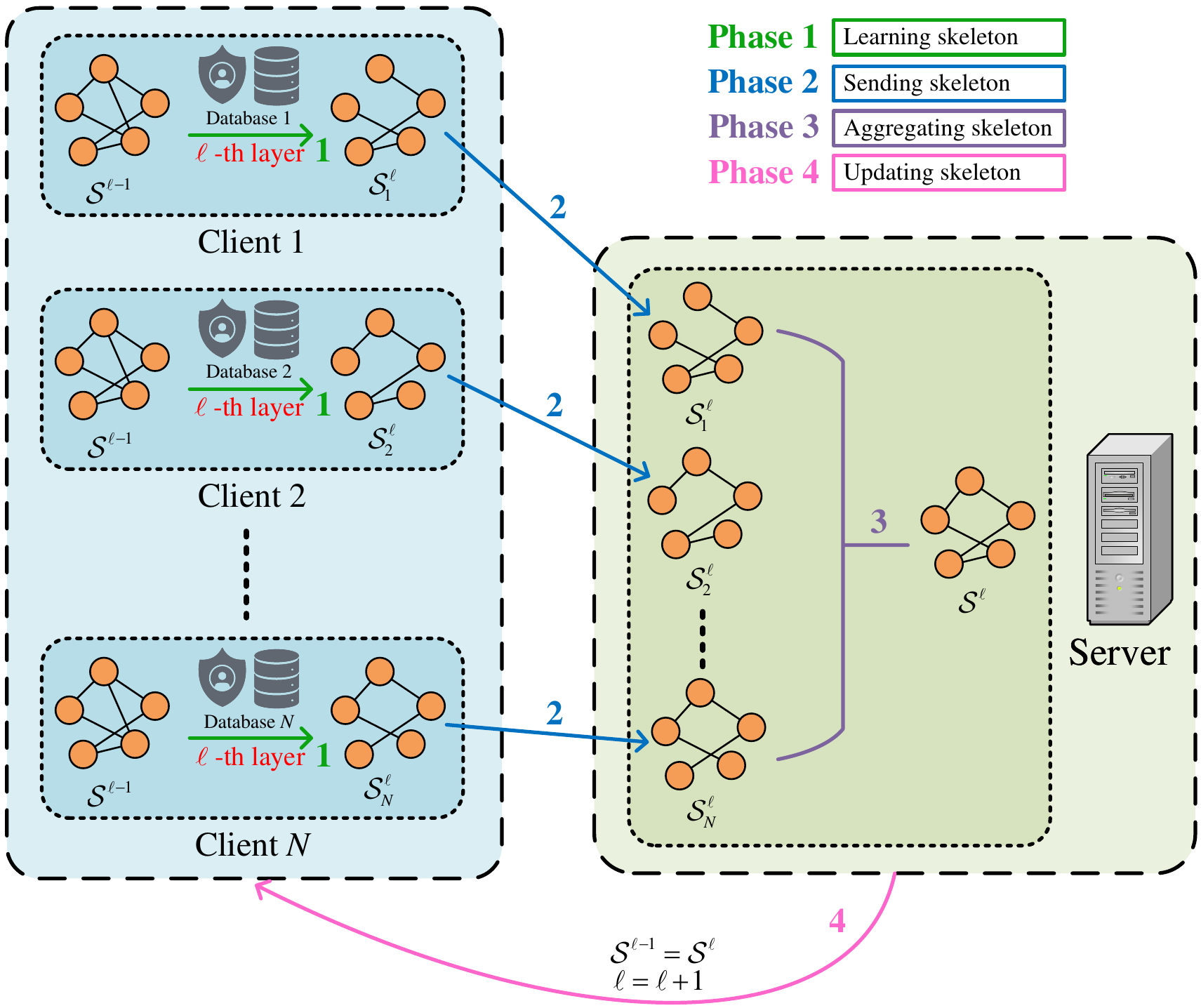}\\
	\caption{The FedSkele subroutine of FedPC}
	\label{FedPC-skeleton}
\end{figure}

\subsubsection{The FedSkele subroutine.}\label{sec3-2-1}

Assuming that there are $N$ clients (labeled as Client 1, Client 2,$\cdots$, Client $N$) and one central server. Inspired by the layer-by-layer skeleton learning idea of the PC algorithm, FedSkele is equipped with a novel layer-wise aggregation strategy to iteratively learn the skeleton with four phases in each layer as shown in Figure~\ref{FedPC-skeleton}.

FedSkele starts from the layer of $\ell=0$ with a complete undirected graph $\mathcal{S}^c$.

\textbf{Phase 1:} 
Each client learns the skeleton at the $\ell$-th layer independently. At the $\ell$-th layer, each client uses the FedSkele subroutine and its local dataset to update the initial skeleton ($\mathcal{S}^{c}$ when $\ell=0$, or the skeleton $\mathcal{S}^{\ell-1}$ obtained at the ($\ell-1$)-th layer) by first setting the size of the conditioning set for CI tests to $\ell$. Then it traverses each adjacent variable pairs $(X_i,X_j)$ in $\mathcal{S}^{\ell-1}$ (or $\mathcal{S}^{c}$ if $\ell=0$), and checks if $\exists\textit{\textbf{Z}}\subseteq \textbf{\textit{ne}}(\mathcal{S}^{\ell-1},X_i)\setminus\{X_j\}$ (or $\exists\textit{\textbf{Z}}\subseteq\textbf{\textit{ne}}(\mathcal{S}^{c},X_i)\setminus\{X_j\}$ if $\ell=0$)  with the size of $\ell$ makes $X_i$ and $X_j$ independent. If so, the edge between $X_i$ and $X_j$ is removed from $\mathcal{S}^{\ell-1}$ (or $\mathcal{S}^{c}$). The skeleton learnt at the end of this phase at client $n$ is $\mathcal{S}^{\ell}_n$ ($n\in\{1, 2, \ldots, N\}$).

\textbf{Phase 2:} Each client sends the skeleton learnt at the end of Phase 1, i.e., $\mathcal{S}^{\ell}_n$ ($n\in\{1, 2, \ldots, N\}$) simultaneously to the server at the $\ell$-th layer.

\textbf{Phase 3:} The server aggregates all skeletons learnt at the $\ell$-th layer. The server receives the learnt skeletons from all clients, then aggregates these skeletons by merging $\mathcal{S}^{\ell}_1$, $\mathcal{S}^{\ell}_2$,$\cdots$, $\mathcal{S}^{\ell}_N$ into a global $\mathcal{S}^{\ell}$. 
Specifically, for each pair of variables $X_i$ and $X_j$, if more than 30\% (a default value) of clients compute that there is an edge between $X_i$ and $X_j$ at the $\ell$-th layer, the edge is kept in $\mathcal{S}^{\ell}$.
In Section~\ref{sec4-5}, we will discuss why we use 30\% as the default threshold.

\textbf{Phase 4:} 
The server sends the aggregated skeleton $\mathcal{S}^{\ell}$ to each client as the initial skeleton for skeleton learning at the ($\ell+1$)-th layer, if the value of $\ell$ is smaller than the maximum number of direct neighbors that a variable has in the $\ell$-th layer skeletons learnt by all clients. Otherwise, the FedSkele subroutine is finished and the FedOrien subroutine starts.

This layer-wise aggregation strategy enables each client only to send and update its skeleton learnt at each layer at the server while protects the data privacy of each client. Then at each layer, by Phases 3 and 4, each client can achieve a high quality of an initial skeleton from the server for skeleton learning especially for the client owning low-quality data. For example, due to a small-sized sample problem, at the  $\ell$-th  layer, if a true edge is wrongly deleted and not in $\mathcal{S}_n^{\ell}$ at Client $n$, by the aggregation strategy, the edge could be added to the skeleton $\mathcal{S}^{\ell}$ for the $(\ell+1)$-th  layer skeleton learning of Client $n$. Thus this layer-wise strategy alleviates the problem of data quality.
Additionally, as we discussed above, through the number of direct neighbors of a variable in the skeleton to control the layer size $\ell$, the strategy guarantees that the FedSkele subroutine can converge to a high-quality skeleton without requiring any special parameters.

\begin{figure}[t]
	\centering
	\includegraphics[width=0.42\textwidth]{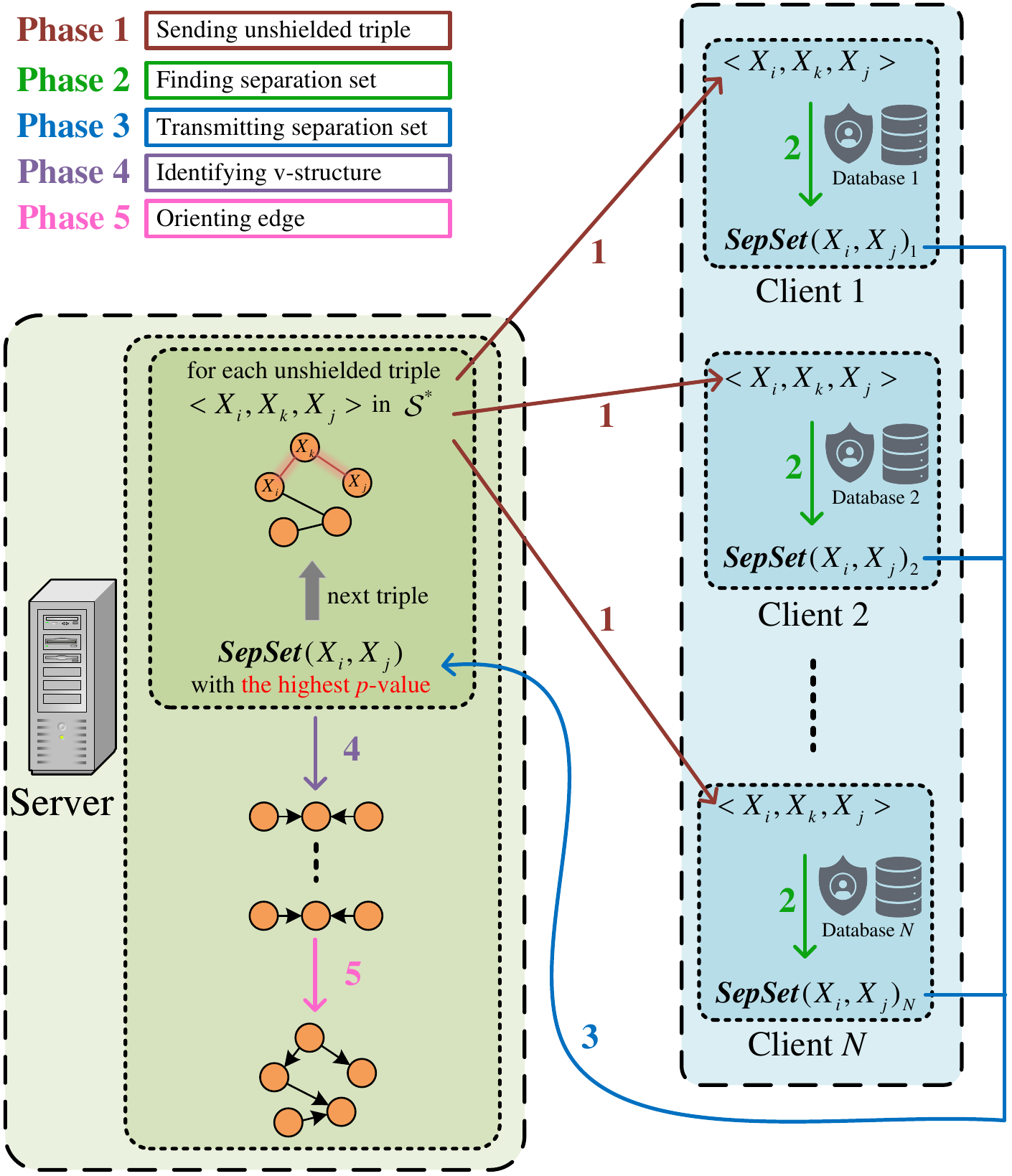}\\
	\caption{The FedOrien subroutine of FedPC}
	\label{FedPC-orientation}
\end{figure}

\subsubsection{The FedOrien subroutine.}\label{sec3-2-2}

Let $\mathcal{S}^*$ denote the final skeleton obtained in the FedSkele subroutine. The FedOrien subroutine first identifies v-structures in $\mathcal{S}^*$, then orients the remaining undirected edges. Note that our proposed FedPC algorithm uses CI tests to learn the skeleton and orient the undirected edges, hence the output of FedPC is a CPDAG instead of a DAG. After five phases in the FedSkele subroutine, we employ the acyclic constraint technology proposed in reference\cite{spirtes2000causation} to transform CPDAG returned by FedPC into DAG.

For an unshielded triple $<X_i,X_k,X_j>$ in $\mathcal{S}^*$, to check whether the triple is a v-structure, as discussed above, we need to have the separation set of $X_i$ and $X_j$  which makes $X_i$ and $X_j$ conditionally independent.
In fact, we can attain such a separation set in the FedSkele subroutine. However, due to potential data quality problems at clients, different clients may have different separation sets for an unshielded triple $<X_i, X_k, X_j>$ in $\mathcal{S}^*$. 
In addition, $\mathcal{S}^*$ is an aggregated skeleton, then the separation set for an unshielded triple $<X_i, X_k, X_j>$ in $\mathcal{S}^*$ may be different from those separation sets attained at the clients in the FedSkele subroutine. Therefore, FedPC does not save separation sets in the FedSkele subroutine. But this brings challenges for finding a separation set for the unshielded triple $<X_i, X_k, X_j>$ that is consistent with $\mathcal{S}^*$ for v-structure learning.

Since each client does not share its raw data with the server, the FedOrien subroutine cannot directly compute the separation set for any non-adjacent variables in $\mathcal{S}^*$ at the server. Thus we have designed an effective strategy for the FedOrien subroutine to identify consistent separation sets for learning v-structures in $\mathcal{S}^*$ across clients. 

As outlined in Figure~\ref{FedPC-orientation}, the FedOrien subroutine consists of the following five phases.

\textbf{Phase 1 (Sending unshielded triples):} 
The server first identifies all the unshielded triples in $\mathcal{S}^*$, then sends each of them and the set of their direct neighors to each client for learning separation sets. Here we use an unshielded triple $<X_i,X_k,X_j>$ in $\mathcal{S}^*$ as an example. During this phase, the server sends the triple $<X_i,X_k,X_j>$ and $\textbf{\textit{ne}}(\mathcal{S}^*,X_i)$ to each client. Then the FedOrien subroutine asks each client to independently find a subset $\textit{\textbf{Z}}\subseteq \textbf{\textit{ne}}(\mathcal{S}^*,X_i)\setminus\{X_j\}$ that makes 
$X_i\indep X_j|\textit{\textbf{Z}}$ hold using its local data.

\textbf{Phase 2 (Finding separation sets):}
Assume the null hypothesis of ``$H_0: X_i \indep X_j|\textbf{\textit{Z}}$", for a CI test of $X_i$ and $X_j$ given a subset \textit{\textbf{Z}}, $X_i \indep X_j|\textbf{\textit{Z}}$ holds, if and only if the $p$-value is greater than $\alpha$ ($\alpha$ is the significance level of the statistical test). At Phase 2, if a client finds that $X_i\indep X_j|\textit{\textbf{Z}}$ holds using its local data, the client  saves the separation set $\textit{\textbf{Z}}$ and the $p$-value of the CI test . When finding all separation sets, the client selects the separation set with the highest p-value as the consistent separation set and send it to the server.


\textbf{Phase 3 (Transmitting separation set):} To achieve a consistent separation set for identifying whether the unshielded triple $<X_i,X_k,X_j>$ in $\mathcal{S}^*$ is a v-structure, all clients send  separation sets and  $p$-values about the unshielded triple $<X_i,X_k,X_j>$ to the server. Then at the sever the FedOrien subroutine selects the separation set with the highest $p$-value from those separation sets as the consistent separation set for v-structure learning. 
The rationale behind the idea is described as follows. For an unshielded triple $<X_i,X_k,X_j>$ in $\mathcal{S}^*$, $X_i$ and $X_j$ are assumed that they are independent in the skeleton $S^*$. Thus, if a CI test of $X_i$ and $X_j$ given a separation set with the highest $p$-value, the separation set has a high probability  to make $X_i$ and $X_j$ really independent and to be a true separation set in the underlying DAG. Then this kind of separation sets can help us identify accurate v structures.

\textbf{Phase 4 (Identifying v-structure):} 
Based on the selected consistent separation sets for all unshielded triples in $\mathcal{S}^*$, at the server, the FedOrien algorithm identifies the v-structures from all unshielded triples without access to each client's data. For example, for the unshielded triple $<X_i,X_k,X_j>$ in $\mathcal{S}^*$, if $X_k$ is not in the separation set of $X_i$ and $X_j$ found in Phase 3, the FedOrien subroutine considers $<X_i,X_k,X_j>$ as a v-structure and orients it as $X_i\rightarrow X_k\leftarrow X_j$.

\textbf{Phase 5 (Orienting remaining edges):} Based on the learnt v-structures, at the server, the FedOrien subroutine orients the remaining undirected edges as many as possible using the Meek rules without requiring clients' data.

\subsection{Privacy preservation and communication cost}\label{sec3-3}

\textbf{\ \ \ \ \ Data privacy when using FedPC.}
In a federated learning setting, when the clients and the sever exchange causal skeletons, it would reveal the independence/dependence relationships between variables. To protect the semantic information of variables while avoiding direct communication between clients, we design an easily implementable privacy protection strategy in the FedPC algorithm. Specifically, we require the remote server to send instructions to each client, requesting them to sort and assign unique identifiers (e.g., ``1", ``2", ``3", ...) to the semantic information of all variables based on their alphabetical order. If variables share the same first letter, they are further sorted based on the second letter of their semantic information, and so on. Each client then sends only the assigned identifiers to the remote server for aggregation (we provide an illustrative example of this strategy in Section S-2 of the Supplementary Material.). Given our assumption that the feature space dimensions are the same across different clients, this strategy ensures variable alignment without the need for communication between clients, while maintaining the confidentiality of variable semantics from the remote server. Meanwhile, compared with existing federated causal structure learning algorithms, FedPC exchanges the skeleton without including edge directions, which further reduces the possibility of leaking sensitive information indirectly.

\textbf{Communication costs of FedPC.}
The communication costs of FedPC are analyzed as follows. 
In the FedSkele subroutine, at each layer, each client needs to send to and receive from the server an adjacency matrix respectively. Since FedPC uses the layer-wise aggregation strategy to learn skeletons, it seems that there might have many communication rounds between the server and clients. 
In fact, the communication costs can be managed and regulated by the maximum size of direct neighbors of a variable in the currently learnt skeleton.
As we discussed in Section \ref{sec4-4}, as the value of  $\ell$ increases, falsely direct neighbors will be removed from the currently learnt skeleton and the sizes of direct neighbors of the variables become small. The  FedSkele subroutine ends if the value of $\ell$ is bigger than the sizes of direct neighbors of all variables in the currently learnt skeleton. Thus, the convergence of this subroutine is determined by the maximum size of direct neighbors of a variable in the learnt skeleton. The subroutine often converges with a small value of $\ell$, especially with a spare underlying DAG. On average, the parameter converges when the value of $\ell$ lies between 3 and 5.
This makes the communication costs of FedPC predictable and adjustable, and the system can be scaled up or down depending on the communication resources available.

\section{Experiments}\label{sec4}

In this section, we conduct experiments to verify the effectiveness of FedPC against its rivals. We first describe the experimental settings in Section~\ref{sec4-1}, then discuss the experimental results on synthetic and real datasets in Sections~\ref{sec4-2} and~\ref{sec4-3}, respectively. Finally, we analyze the parameters related to FedPC in~Section~\ref{sec4-4} and~\ref{sec4-5}.

\subsection{Experiment settings}\label{sec4-1} 
\subsubsection{Datasets.}\label{sec4-1-1} 
\begin{table}
	\caption{Details of the five benchmark Bayesian networks}
	\centering
	\begin{tabular}{c c c c}
		\toprule
		Network & \tabincell{c}{Number of\\variables} & \tabincell{c}{Number of\\edges} & \tabincell{c}{Maximum\\in/out-degree} \\
		\midrule
		alarm & 37 & 46 & 4/5 \\
		insurance & 27 & 52 & 3/7 \\
		win95pts & 76 & 112 & 7/10 \\
		andes & 223 & 338 & 6/12 \\
		pigs & 441 & 592 & 2/39 \\
		\bottomrule
	\end{tabular}\label{data-detail}
\end{table}

The datasets used in the experiments include the following synthetic and real datasets. 

We assume that there are $K$ data samples in a dataset and $N$ clients exists, and $N$ lies in $\{3, 5, 10, 15\}$. To introduce unevenness, we randomly assigned the data samples to each client while ensuring that each client contains at least $\frac{K}{{N*2}}$ data samples. This approach aims to ensure that the data distribution is not heavily skewed towards any specific client. The dataset in each client has the same set of variables and we assume that the dataset in each client is generated from the same ground-truth DAG.

\begin{itemize}

	\item \textbf{Benchmark Bayesian network (BN) datasets.}
	We use five benchmark BNs, alarm, insurance, win95pts, andes and pigs, to generate five discrete datasets, respectively. Each dataset contained 5000 samples. The details of the five benchmark BNs are presented in Table~\ref{data-detail}.
	
	\item \textbf{Synthetic datasets.} 
	We conduct FedPC on synthetic linear and nonlinear datasets. We present experimental results of FedPC on linear datasets in Section~\ref{sec4-2}, and the results on nonlinear datasets in Section S-1-3 in the Supplementary Material.
	For the linear synthetic datasets, we generate five continuous datasets using an open-source toolkit \cite{kalainathan2020causal}  with the number of variables to 10, 20, 50, 100 and 300 respectively. Each synthetic dataset contains 5000 continuous samples. The generative process employs a linear causal mechanism represented as follows:
	\begin{equation}
		y = \textbf{X}W+\times E,
		\label{eq-0}
	\end{equation}
	where $+\times$ denotes either addition or multiplication, \textbf{X} denotes the vector of causes, and $E$ represents the noise variable accounting for all unobserved variables.
	For the nonlinear synthetic datasets, the causal mechanism used in the generative process is Gaussian Process (GP), and the mechanisms are represented as: 
	\begin{equation}
		y = GP(\textbf{X})+\times E.
		\label{eq-0}
	\end{equation}
	
	The proportion of noise in the mechanisms is set to 0.4. Gaussian noise was used in the generative process. On nonlinear datasets, we utilize the Kernel-based Conditional Independence test (KCI-test) \cite{zhang2012kernel} instead of Fisher's Z Conditional Independence test for detecting nonlinear dependencies. 
	
	\item \textbf{Real dataset.} We also compare FedPC with its rivals on a real biological dataset with 853 samples, Sachs~\cite{sachs2005causal}. Sachs is a protein signaling network expressing the level of different proteins and phospholipids in human cells. It is commonly considered as a benchmark graphical model with 11 nodes (cell types) and 17 edges.
	
\end{itemize}

\subsubsection{Evaluation metrics.}\label{sec4-1-2} 
To evaluate the performance of FedPC with its rivals, we use the following frequently used metrics in causal structure learning (More  evaluation metrics and the corresponding results please see the Supplementary Material).

\begin{itemize}
	\item SHD (Structural Hamming Distance). The value of SHD is calculated by comparing the learnt causal structure with the true causal structure. Specifically, the value of SHD is the sum of undirected edges, reverse edges, missing edges and extra edges. 
	In Section S-1 of the Supplementary Material, we present reverse, extra, miss metrics of our method and its rivals to further validate the effectiveness of our method.
	
	\item F1. The F1 measure is a comprehensive evaluation metric, and it is calculated as $F1 = \frac{{2*Recall*Precision}}{{Recall+Precision}}$.  Precision is equal to the number of correctly predicted arrowheads in the output divided by the number of edges in the output of an algorithm, while Recall is the number of correctly predicted arrowheads in the output divided by the number of true arrowheads in the true causal structure. We present more experimental results in Section S-1 of the Supplementary Material, where Precision and Recall are shown to compare the performance of FedPC with that of its rivals.
	SHD and F1 are used to measure structure error and structure correctness, respectively.
	\item Time. We report running time (in seconds) as the efficiency measure of different algorithms.
\end{itemize}

FedPC utilizes the PC algorithm to learn the skeleton and orient the undirected edges in the causal graph. Therefore, the output of FedPC is a CPDAG, which contains both directed and undirected edges. We employ the acyclic constraint technology proposed in reference\cite{spirtes2000causation} to transform CPDAGs returned by FedPC into DAGs, and then calculated the Structural Hamming Distance (SHD) and F1 scores of these DAGs.

\subsubsection{Comparison methods.}\label{sec4-1-3}

FedPC is compared with 8 baselines as follows.

(1) NOTEARS-Avg. We run the NOTEARS~\cite{zheng2018dags} algorithm at each client independently, then calculate the averaging results of SHD and F1 of all learnt DAGs as the final results. 

(2) NOTEARS-ADMM. We run the NOTEARS-ADMM algorithm~\cite{ng2022towards}, and then calculate the SHD and F1 of the learnt DAG as the final results.

(3) FedDAG. We run the FedDAG~\cite{gao2023feddag} algorithm and then calculate the SHD and F1 of the learnt DAG as the final results.

(4) PC-Avg. We first run the PC algorithm at each client independently for obtaining $N$ DAGs ($N$ is the number of clients), and then calculate the averaging SHD values and F1 values of all learnt DAGs as the final results of PC-Avg. 

(5) PC-Best. We first run the PC algorithm at each client independently to get $N$ DAGs, and then select the DAG with the lowest SHD value as the final output.

(6) PC-All. We centralize all clients' data to a single dataset and run the PC algorithm on it.

(7) FedPC-Simple-I. We run the PC algorithm at each client independently to learn the DAGs, then aggregate all learnt DAGs at the server by the strategy that if more than 30\% (the same ratio as our method) of the learnt DAGs contain a directed edge between two variables, this edge is kept in the final DAG.

(8) FedPC-Simple-II. We run the PC algorithm to learn the skeletons independently at each client, then aggregate all learnt skeletons at the server by the strategy that if an undirected edge between two variables exists on more than 30\% of the skeletons, this edge will be kept in the final skeleton.
Then we take the intersection of the separation sets between two variables learnt from each client to learn v-structures.
This is an ablation study of our proposed algorithm, by removing the layer-wise strategy of the FedOrien subroutine.
\subsubsection{Implementation Details.}\label{sec4-1-4} 
All experiments were conducted on a computer with Intel Core i9-10900F 2.80-GHz CPU and 32-GB memory. The significance level for CI tests is set to 0.01. For PC, NOTEARS, and NOTEARS-ADMM, we used the source codes provided by their authors. NOTEARS-Avg and NOTEARS-ADMM use 0.3 as the threshold to prune edges in a DAG and FedDAG uses 0.5 as the threshold, those are the same as the original paper. The source  codes of FedPC are provided in the Supplementary Materials.

\subsection{Results on synthetic and benchmark data}\label{sec4-2}

In the section, we report SHD, F1 and running time of FedPC and its rivals, respectively.

\subsubsection{The SHD metric.}\label{sec4-2-1}

\begin{figure*}
	\centering
	\includegraphics[width=0.94\linewidth]{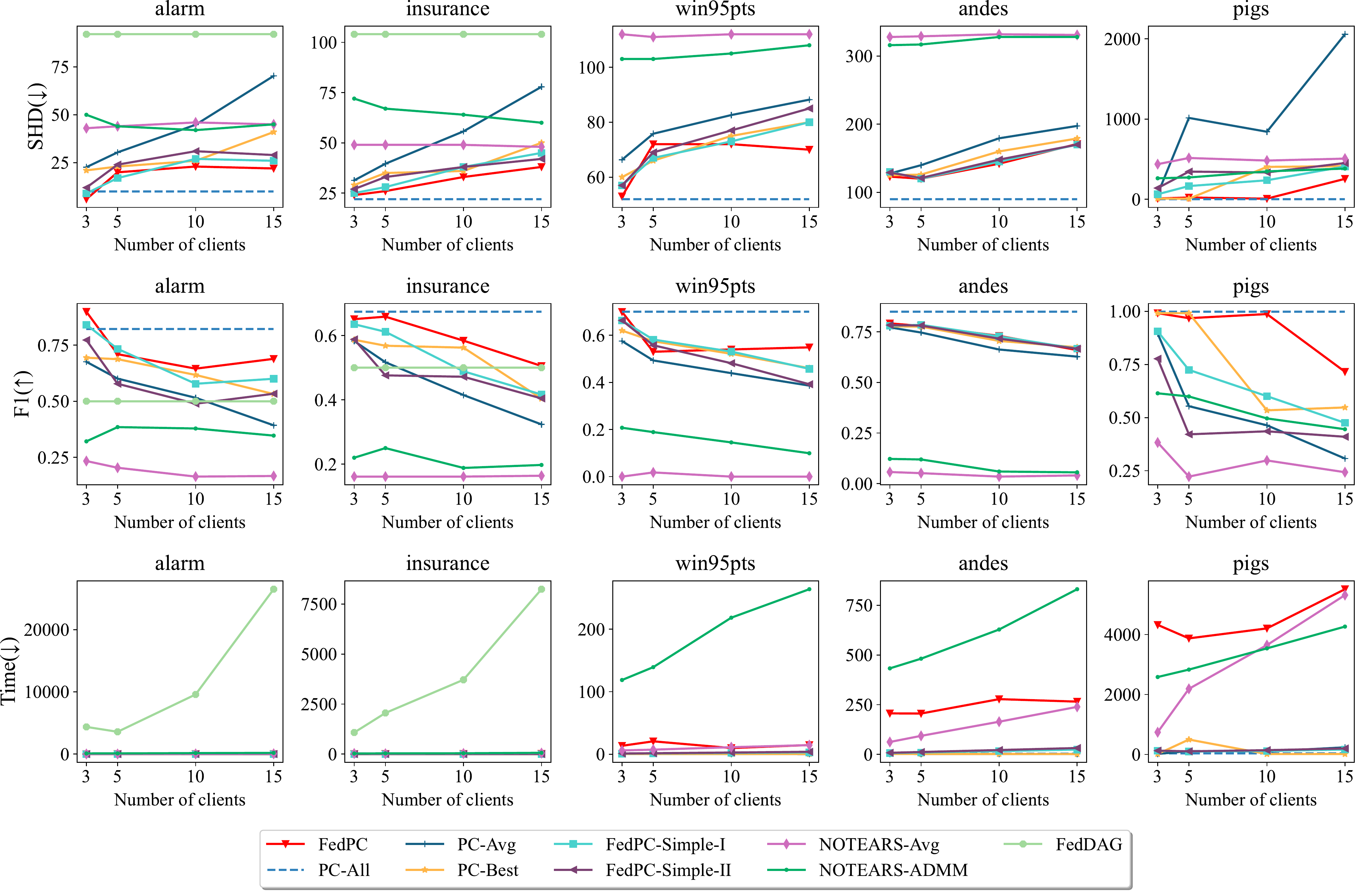}
	\caption{Results on 5 benchmark BN datasets}
	\label{fig-syn-dis}
\end{figure*}

Figures~\ref{fig-syn-dis} to~\ref{fig-syn-con} (the top row in each figure) show the SHD values of FedPC and its 8 rivals using five discrete BN datasets and five continuous synthetic datasets, respectively. The smaller value of SHD denotes better performance of an algorithm. We can see that FedPC outperforms its six rivals. 
\begin{itemize}
	
	\item The value of SHD of FedPC is much smaller than that of NOTEARS-Avg, NOTEARS-ADMM and FedDAG on all datasets. Since it is hard for the two rivals to select a suitable threshold to prune false directed edges, the DAGs learnt by these two rivals often contain a larger number of false edges, leading to inaccurate DAGs.
	
	\item FedPC is superior to FedPC-Simple-I and FedPC-Simple-II on the SHD metric. FedPC-Simple-I directly applies the PC algorithm to each client and directly aggregates the learnt DAGs to get the final DAG. FedPC-Simple-II simply aggregates skeletons and takes the intersection of the separation sets to orient edges.
	However, FedPC uses a layer-wise aggregation strategy for accurate skeleton learning and a  consistent separation set identification strategy for  accurate edge orientations.
	
	\item  FedPC is better than PC-Avg and PC-Best, since PC-Avg and PC-Best do not exchange information between a client and the server. This further verifies the effectiveness of the two strategies of FedPC. The results obtained by FedPC may not be as accurate as those have a sufficient sample size like PC-All because PC-All centralizes all independent clients’ datasets. However, on synthetic datasets, FedPC performs better than PC-All. The explanation is that synthetic datasets is ideal while benchmark networks and the real dataset are not ideal datasets. So data samples with uncertainty cannot be generated in synthetic dataset.
	
	\item We can observe that as the number of clients increases, the Structural Hamming distance (SHD) metric increases while the F1 metric decreases. This is due to the reduced number of samples per client when the total number of data samples among all clients. The decrease in sample size interferes with the accuracy of the conditional independence test and reduces the confidence in accepting the null hypothesis of conditional independence between two variables. Consequently, a larger number of erroneous separation sets are identified, leading to incorrect identification of v-structures and a decline in the algorithm's performance. However, to mitigate the impact of the reduced sample size, we choose the separation set with the highest p-value, indicating the highest likelihood of conditional independence between the two variables. This approach minimizes the influence of the smaller sample size on the final determination of independence or dependence.
	
\end{itemize}

\subsubsection{The F1 metric.}\label{sec4-2-2}
\begin{figure*}
	\centering
	\includegraphics[width=0.94\linewidth]{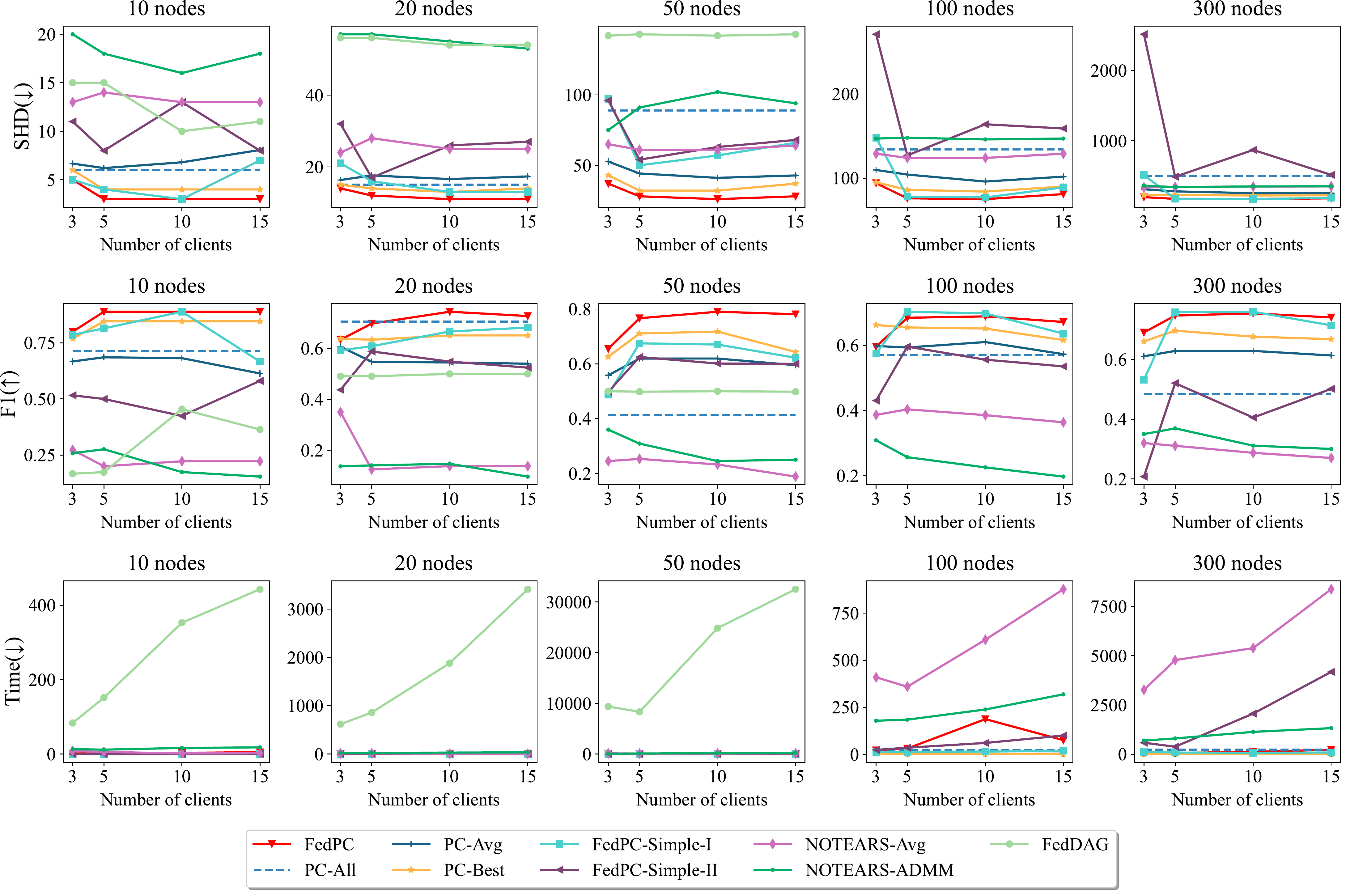}
	\caption{Results of 5 synthetic datasets}
	\label{fig-syn-con}
\end{figure*}

Figures~\ref{fig-syn-dis} to~\ref{fig-syn-con} (the second row in each figure) show the F1 values of FedPC and its rivals using five benchmark BN datasets and five synthetic datasets, respectively. The higher value of F1 denotes better performance of an algorithm. We can see that FedPC achieves a higher F1 than its rivals.
The explanations are as follows. 

FedDAG identifies most of correct directed edges while many wrong edges are included. NOTEARS-Avg and NOTEARS-ADMM identify only a small number of correct directed edges, while PC-Avg and PC-Best wrongly remove many correct edges. FedPC-Simple-I and FedPC-Simple-II just aggregate the DAGs and skeletons learnt by each client and lack an effective strategy to identify consistent separation sets, thus there are many false edges in the final result.

In addition, NOTEARS-Avg and NOTEARS-ADMM are sensitive to the user-defined threshold. Different clients may have different thresholds for edge pruning. An unsuitable threshold may prune either correct edges or retain false edges.

As the number of clients increases, using the five BN datasets, the F1 and SHD values of FedPC and its six rivals decrease, while they do not change much, even increase a little. The possible explanation is that PC-derived algorithms  employ the chi-squared test for discrete data while using the Fisher Z test for continuous data for CI tests. In general, PC-derived algorithms need more data samples for discrete BN data sets than continuous synthetic data for reliable CI tests. Then as the number of clients increases, the number of data samples of each client becomes insufficient, accordingly the performance of FedPC and its six rivals degrades a little using discrete data. However, with FedPC a client owns insufficient data examples leading to incorrect CI tests, the layer-wise aggregation strategy makes FedPC exchange enough information between clients and the server for accurate DAG learning.  

To give a comprehensive performance comparison between FedPC with its rivals, in Section~\ref{sec4-2-4}, we conducted statistical tests to show that FedPC is significantly better than other methods.

\subsubsection{Time Efficiency.}\label{sec4-2-3}
Figures~\ref{fig-syn-dis} to~\ref{fig-syn-con} (the last row in each figure) show the execution time of FedPC and its rivals on the five discrete BN datasets and the five continuous synthetic datasets. 
With the five continuous datasets and most BN datasets, FedPC is faster than NOTEARS-Avg and NOTEARS-ADMM.
FedPC is a little slower than PC-Avg, PC-Best, FedPC-Simple-I and FedPC-Simple-II since FedPC needs more time than those baselines for communications between clients and the server during skeleton learning and it also needs to find separation sets at each clients for getting consistent separation sets.
As the number of clients increases, the running time of most algorithms increases accordingly. FedPC spends more time on the pigs dataset since the size of direct neighbors of variables in the pigs dataset is much larger than that in the other datasets, leading to more time to learn skeletons and separation sets. In summary, at most times, FedPC is competitive with PC-Avg, PC-Best, FedPC-Simple-I and FedPC-Simple-II on running time and faster than NOTEARS-Avg and NOTEARS-ADMM.

\subsubsection{ Statistical Tests.}\label{sec4-2-4}
To give a comprehensive performance comparison between FedPC with its rivals, the Friedman test and Nemenyi test~\cite{demvsar2006statistical} are performed.

We first perform the Friedman test at the 0.05 significance level under the null-hypothesis, which states that the performance of all algorithms is the same on all datasets (i.e., the average ranks of all algorithms are equivalent). Then, we perform the Nemenyi test, which states that the performance of two algorithms is significantly different if the corresponding average ranks differ by at least one critical difference (CD). Note that we only perform these tests on the datasets where all algorithms can run the results.

\begin{figure}[htbp]
	\centering
	\subfigure[F1 metric of each algorithm on synthetic dataset.]{
		\includegraphics[width=0.43\textwidth]{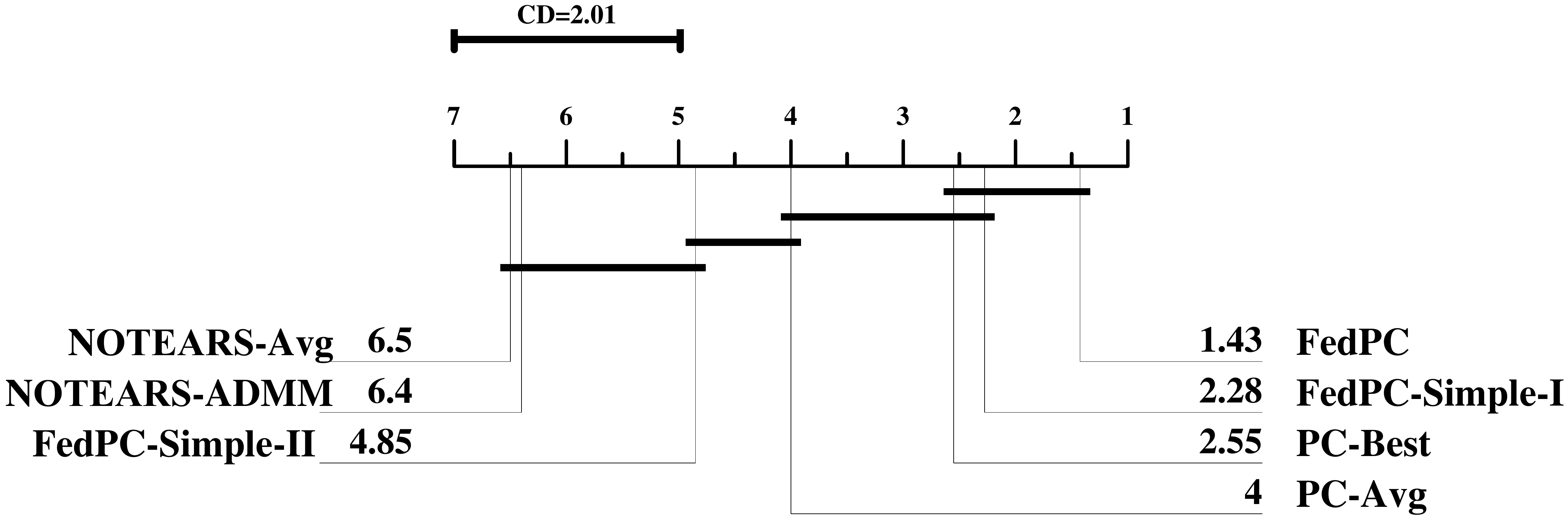}
	}
	\subfigure[SHD metric of each algorithm on synthetic dataset.]{
		\includegraphics[width=0.43\textwidth]{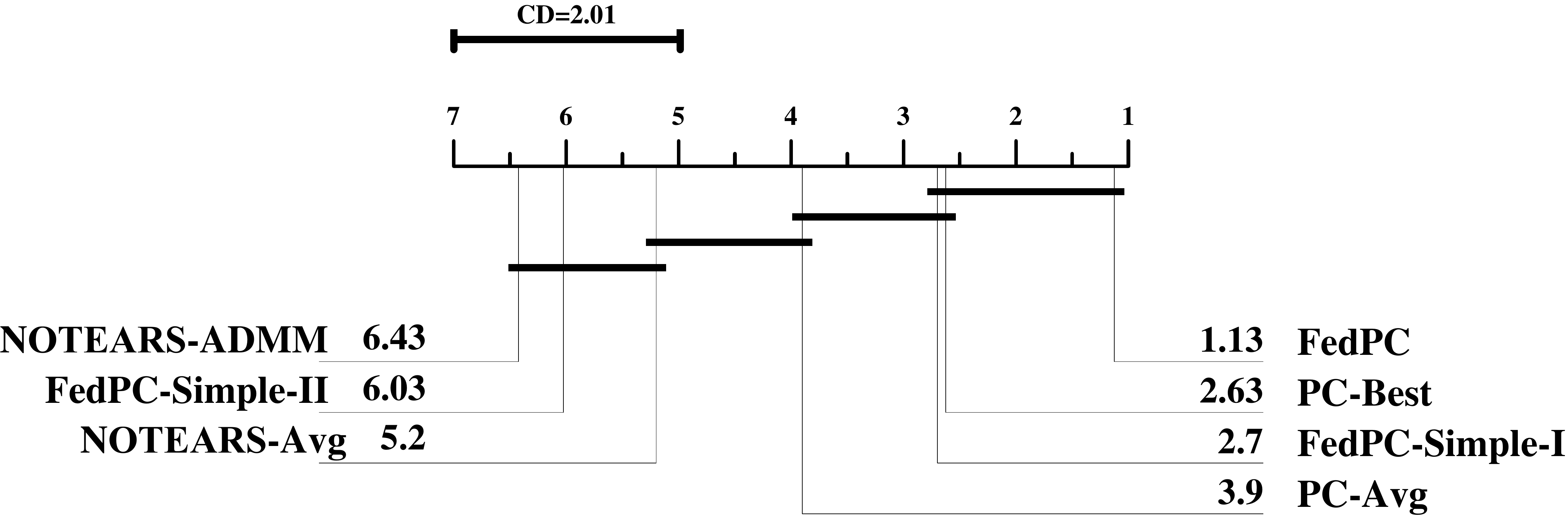}
	}
	\subfigure[F1 metric of each algorithm on benchmark datasets.]{
		\includegraphics[width=0.43\textwidth]{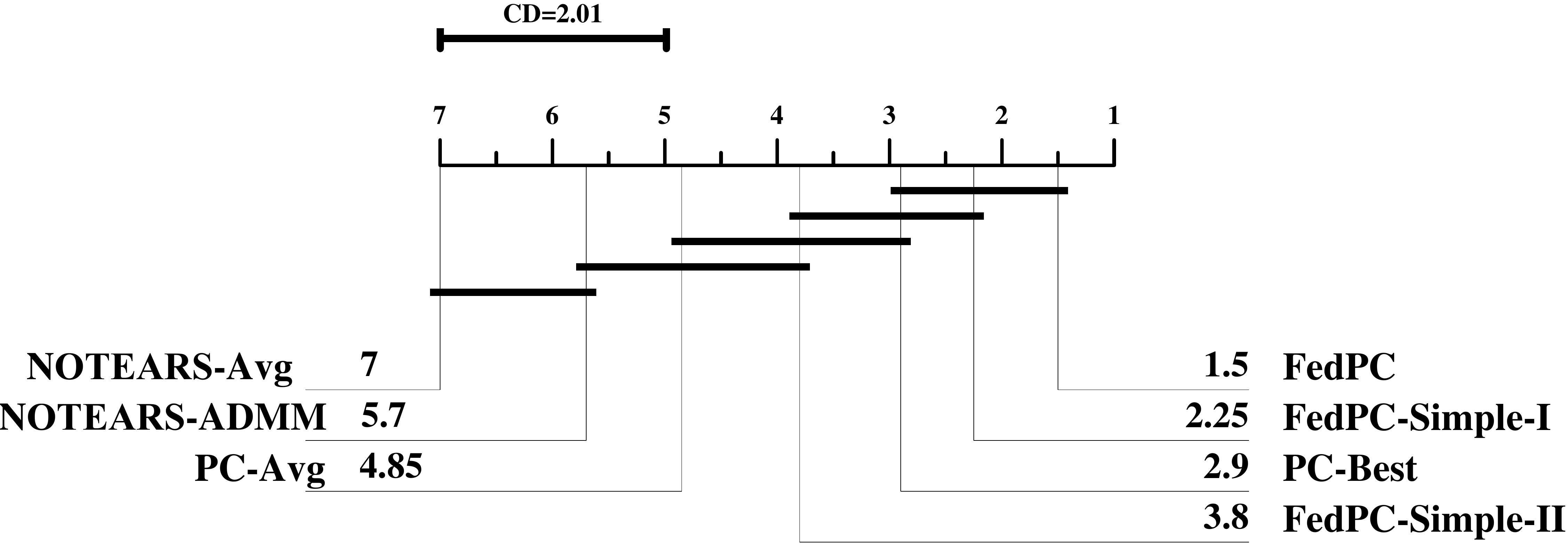}
	}
	\subfigure[SHD metric of each algorithm on benchmark datasets.]{
		\includegraphics[width=0.43\textwidth]{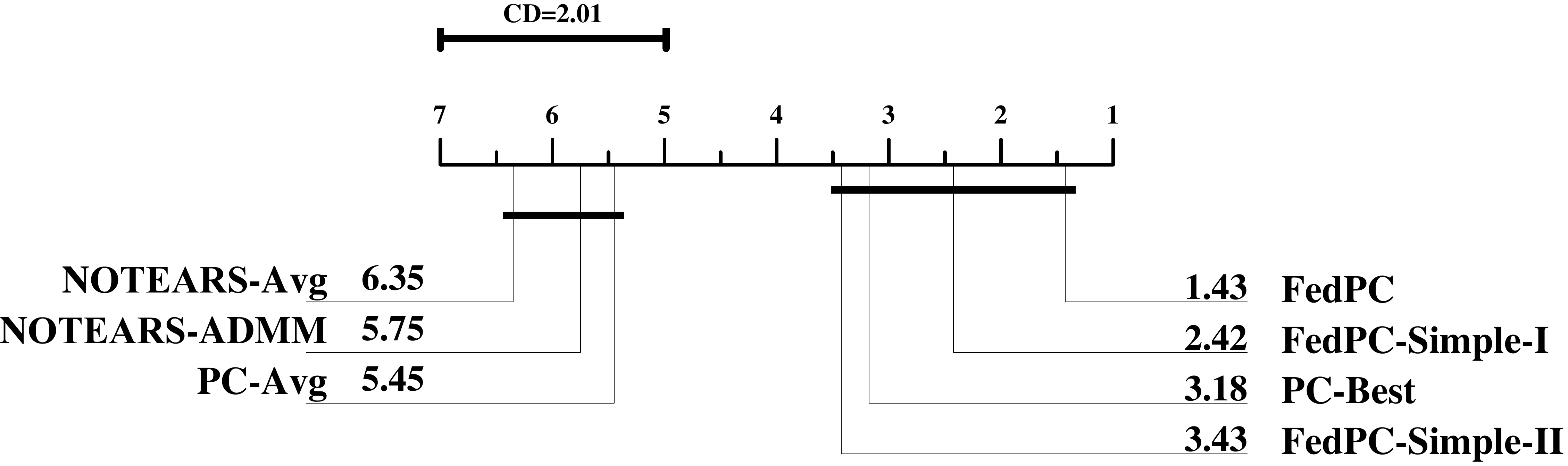}
	}
	\caption{Comparison of FedPC against its rivals with the Nemenyi test. (the lower the rank value, the better the performance.)}
	\label{fig-CD}
\end{figure}

Figs.~\ref{fig-CD}(a), (b), (c) and (d) provide the CD diagrams, where the average rank of each algorithm is marked along the axis (lower ranks to the right). FedPC is the only algorithm that achieves the lowest rank value on both synthetic and benchmark datasets.
On synthetic datasets, whether F1 metric or SHD metric, we observe that FedPC achieves a comparable performance against PC-Best and FedPC-Simple-I, and it is significantly better than the other algorithms.
In terms of the SHD metric of each algorithm on BN datasets, we note that FedPC significantly outperforms NOTEARS-Avg, NOTEARS-ADMM and PC-Avg, and it achieves a comparable performance against the other algorithms.

\subsection{Results on real data}\label{sec4-3}
\begin{figure*}[t]
	\centering
	\includegraphics[width=0.9\linewidth]{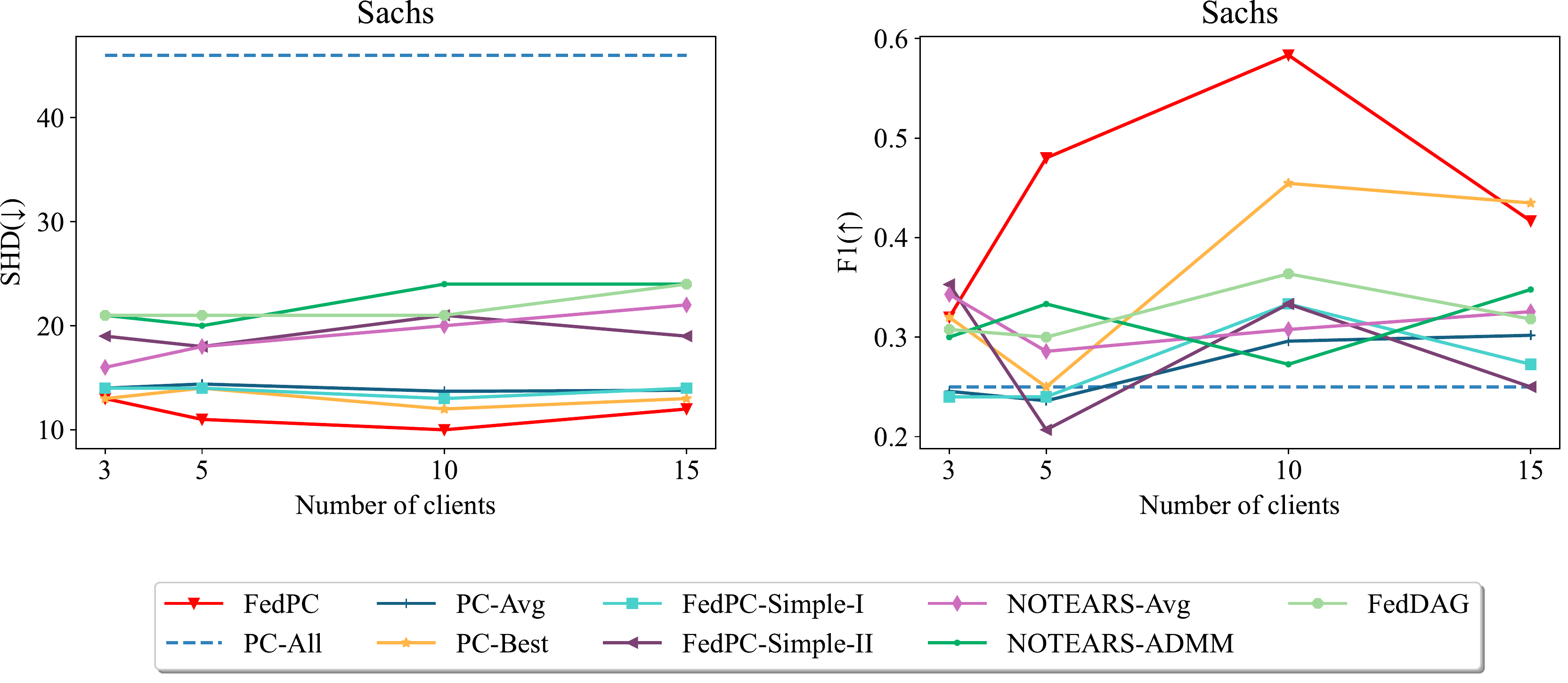}
	\caption{Results of the real dataset}
	\label{fig-real}
\end{figure*}

In this section, we compare FedPC with its rivals on a real dataset, Sachs~\cite{sachs2005causal}, and the experimental results are shown in Figure~\ref{fig-real}. From Figure~\ref{fig-real}, we can see that FedPC achieves the lowest value of SHD regardless of the number of clients. For the F1 metric, when the number of clients is 5 or 10, the F1 value of FedPC is significantly higher than that of other algorithms.
Due to the small scale of the Sachs dataset, the execution time of each algorithm is trival and thus we do not report the running time in Figure~\ref{fig-real}.

\subsection{Parameter analysis}\label{sec4-4}
\begin{figure}[htbp]
	\centering
	\includegraphics[width=0.81\linewidth]{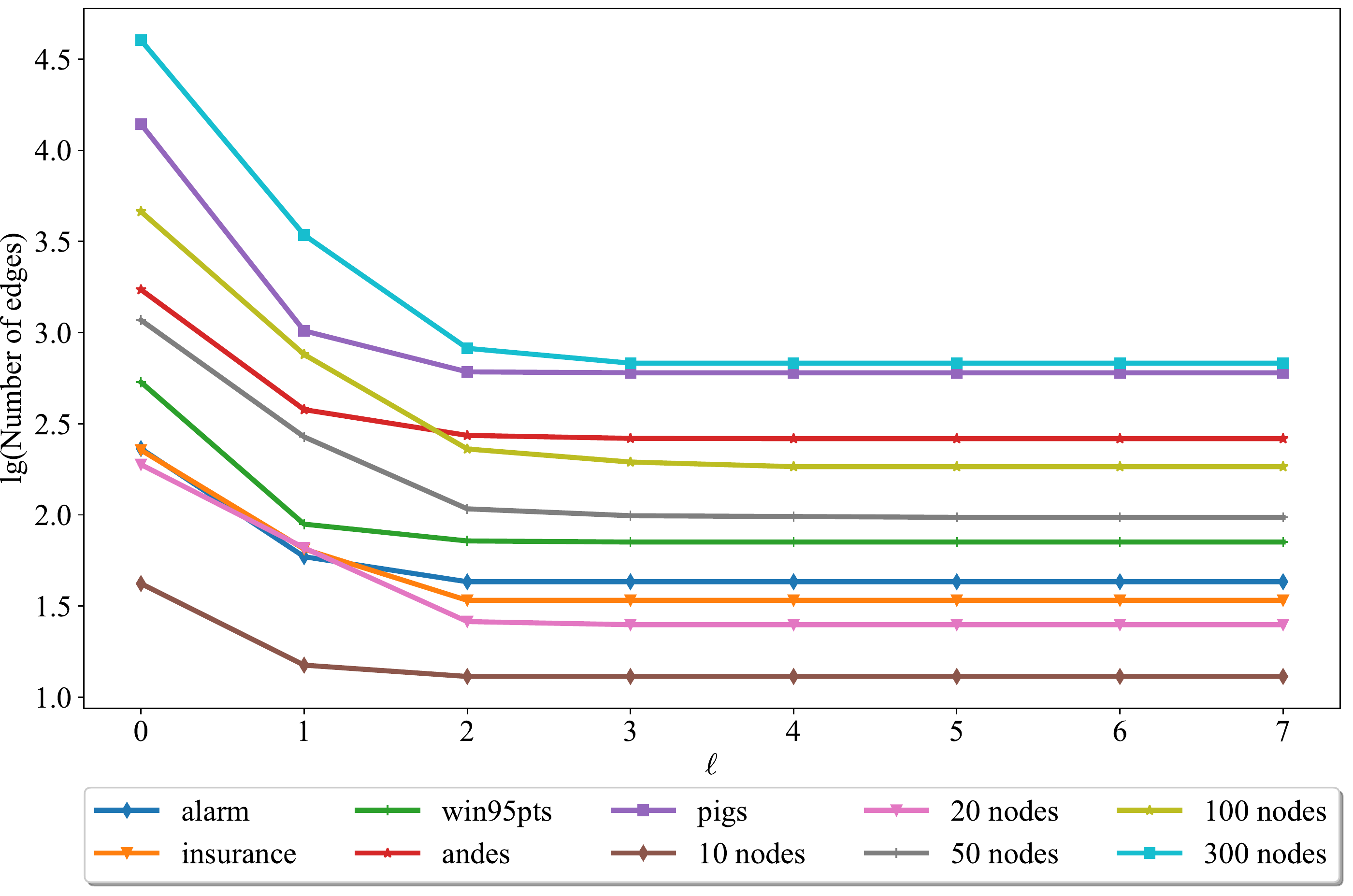}
	\caption{The value of $\ell$ and the convergence of FedPC}
	\label{fig-layer_size}
\end{figure}

As discussed above, the value of $\ell$ is determined by the maximum size of direct neighbors of a variable in the skeleton. 
Figure~\ref{fig-layer_size} shows the relation of the value of $\ell$ and the convergence of FedPC. We can see that the number of edges in a skeleton gradually decreases when $\ell$ is from 0 to 5. On average, the FedSkele subroutine converges when the value of $\ell$ lies between 3 and 5. The possible explanation is that the underlying DAG of a dataset is often a sparse one.

\subsection{Analysis of the ratio of clients.}\label{sec4-5}
%
%

In Phase 3 of the FedSkele subroutine, a threshold ratio of 30\% is chosen to remove or keep an edge in an aggregated skeleton. This threshold ratio is chosen after considering that if the ratio is too low, many incorrect edges may be kept in the skeleton, while if it is too high, many correct edges may be removed from the skeleton.

\begin{figure}[htbp]
	\centering
	\includegraphics[width=0.78\linewidth]{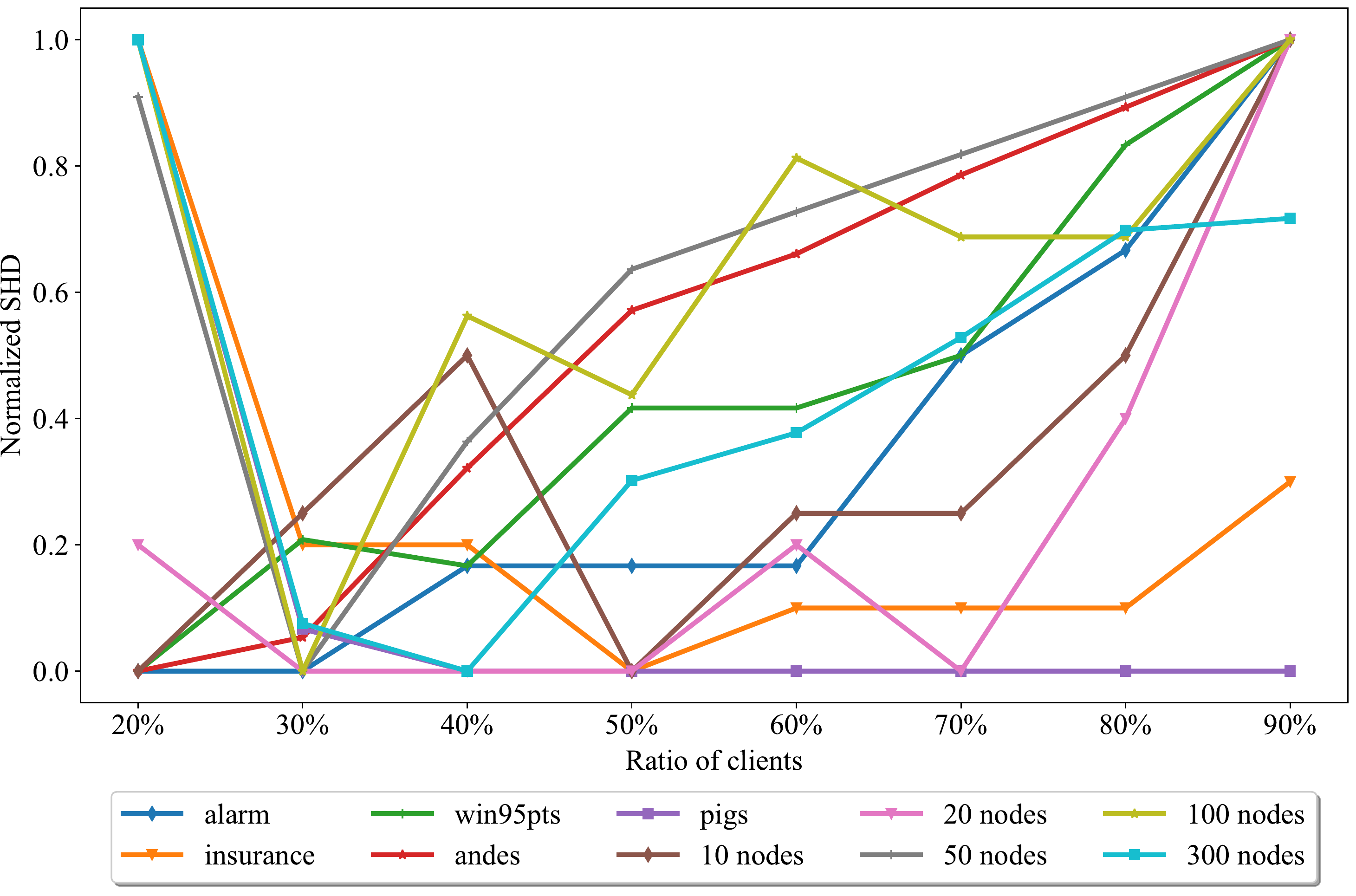}
	\caption{Normalized SHD and the ratio of clients}
	\label{fig-ratioclients}
\end{figure}

To further analyze the effect of the threshold ratio, experiments were conducted with ratios ranging from 20\% to 90\%. The SHD values of the aggregated skeleton using all 10 datasets were observed, and the values were normalized using min-max normalization to scale them to a range of $(0, 1)$. The normalized values of SHD with different ratios are shown in Figure~\ref{fig-ratioclients}.

Empirical results showed that, in general, the SHD value first decreases and then increases as the ratio of clients increases. However, the minimum value of SHD was reached on most of the datasets when the ratio was up to 30\%. This suggests that a threshold ratio of 30\% is an appropriate value for achieving good performance in FedSkele.

\section{Conclusion and Future Work}\label{sec:conclusion}
Learning causal structures while preserving data privacy brings many challenges to traditional causal structure learning methods.
In this paper, we have studied the privacy-aware causal structure learning problem in the federated learning setting and have developed a novel FedPC algorithm that seamlessly have the classic PC algorithm into the federated learning paradigm. Compared to 8 competing algorithms, FedPC achieves  promising performance using various datasets. Furthermore, the idea of FedPC can be applied to existing PC-derived algorithms for designing new  algorithms for tackling data privacy.


Meanwhile, we briefly discuss a few possible research directions for future work. 
Firstly, it is interesting to systematically analyze the different strategies for identifying separation sets for identifying v structures in the FedOrien subroutine. One strategy is to calculate the p-value for all possible separation sets of two variables in an unshielded triple and send them to the server to compute the average p-value (across different clients) for each separation set, then select the separation set with the highest p-value. Another strategy is that we find all separation sets at each client that meets a given threshold, and send these separation sets to the server to perform a voting across these separation sets collected for getting a suitable separation set.
Secondly, in the practical setting, multiple privacy-preserving datasets may contain hidden variables  or heterogeneous, we plan to extend FedPC to deal with either hidden variables or heterogeneous data. Thirdly, we will combine FedPC with IDA~\cite{maathuis2009estimating} for studying privacy-aware cause effect estimation for robust machine learning. Finally, in practical scenarios, the variables may differ across clients. It is worth studying that each client has a different set of variables but all clients share a common set of variables in federated setting.


%

\ifCLASSOPTIONcompsoc

\section*{Acknowledgments}
\else
\section*{Acknowledgment}
\fi

This work was supported by the National Key Research and Development Program of China (under grant
2021ZD0111801) and the National Natural Science Foundation of China (under grant 62176082).

\ifCLASSOPTIONcaptionsoff
\newpage
\fi


%

\bibliographystyle{IEEEtran}
\bibliography{reference}
%
%

%
\vspace{-25pt}
\begin{IEEEbiography}[{\includegraphics[width=1in,height=1.25in,clip,keepaspectratio]{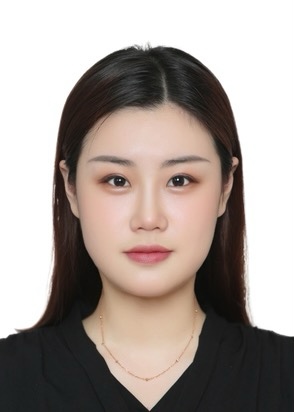}}]
	{Jianli Huang} received the B.S. degree in computer science from Jiangsu University, Zhenjiang, China, in 2021. She is currently pursuing the master’s degree with the School of Computer Science and Information Engineering, Hefei University of Technology, Hefei. 
	
	Her current research interests focus on causal discovery and federated learning.
\end{IEEEbiography}
\vspace{-30pt}
\begin{IEEEbiography}[{\includegraphics[width=1in,height=1.25in,clip,keepaspectratio]{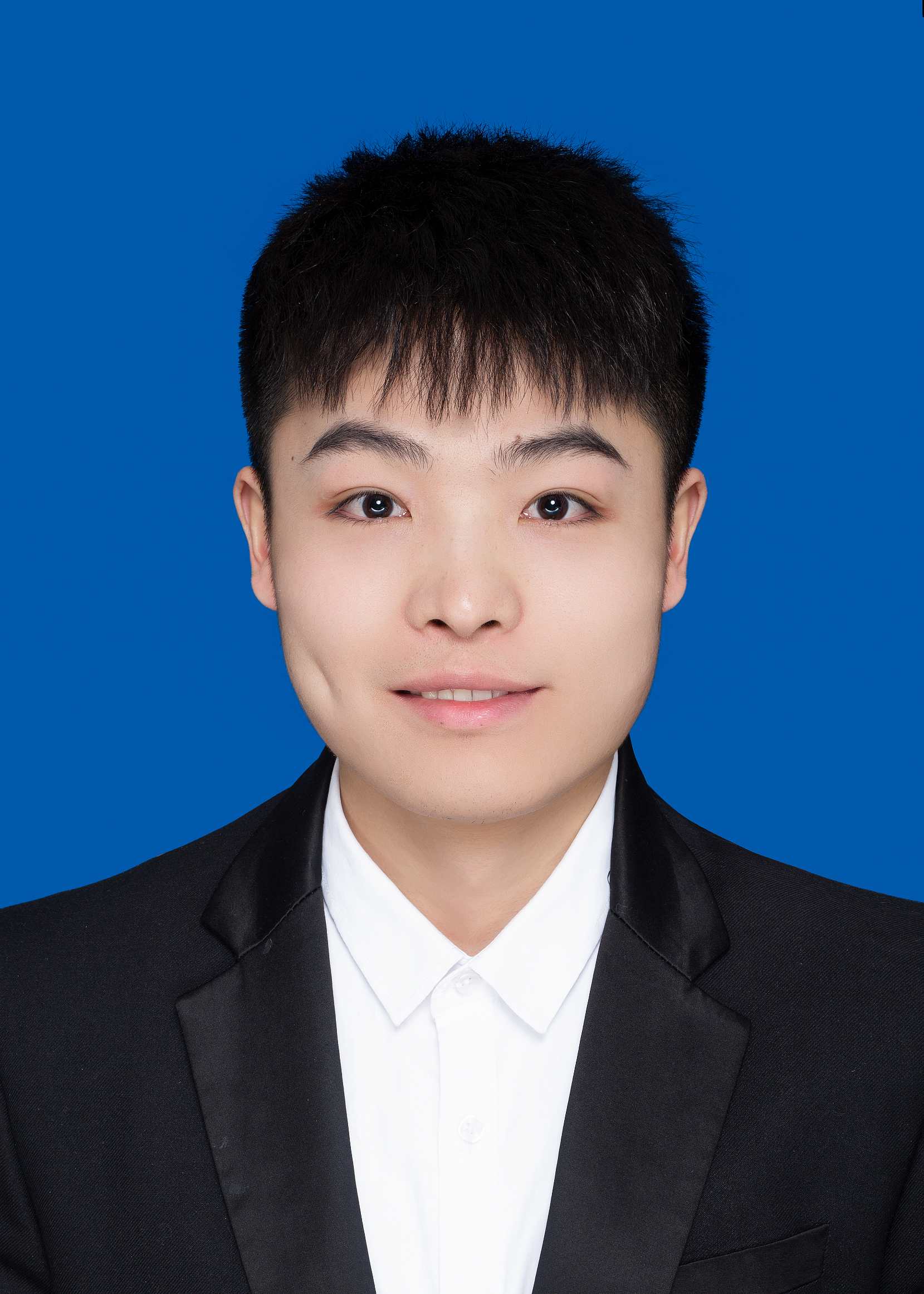}}]
	{Xianjie Guo} received the B.S. degree from Anhui Normal University, Wuhu, China, in 2018. He is currently pursuing the Ph.D. degree with the School of Computer Science and Information Engineering, Hefei University of Technology, Hefei, China.
	
	His current research interests include causal discovery and federated learning.
\end{IEEEbiography}
\vspace{-30pt}
\begin{IEEEbiography}[{\includegraphics[width=1in,height=1.25in,clip,keepaspectratio]{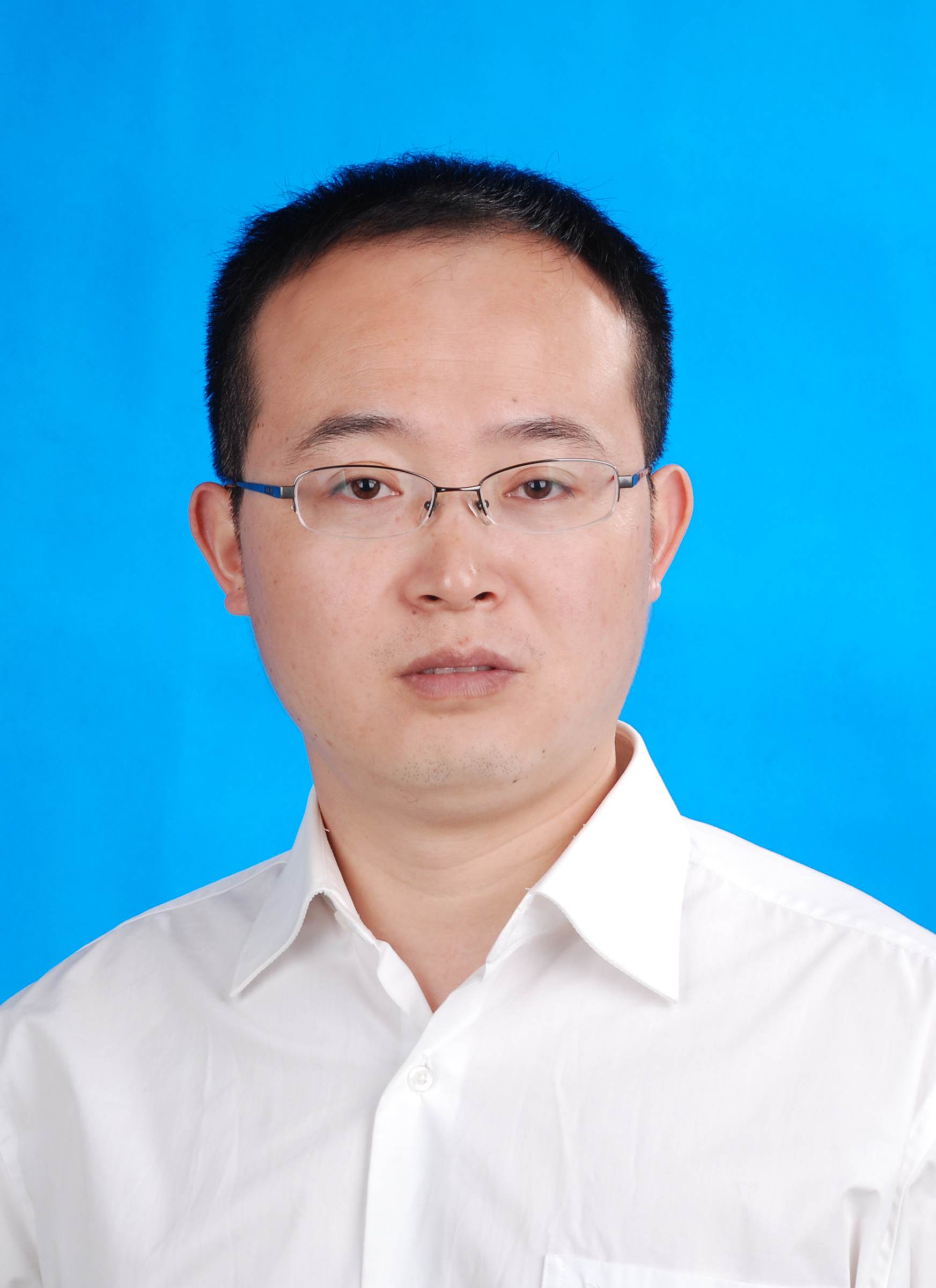}}]
	{Kui Yu} (Member, IEEE) received the Ph.D. degree in computer science from the Hefei University of Technology, Hefei, China, in 2013.
	
	From 2013 to 2018, he was a research fellow of computer science with the University of South Australia, Adelaide, Australia and Simon Fraser University, Burnaby, Canada. He is a full Professor with the School of Computer Science and Information Engineering, Hefei University of Technology. His main research interests include causal discovery and machine learning.
\end{IEEEbiography}
\vspace{-30pt}

\begin{IEEEbiography}[{\includegraphics[width=1in,height=1.2in,clip,keepaspectratio]{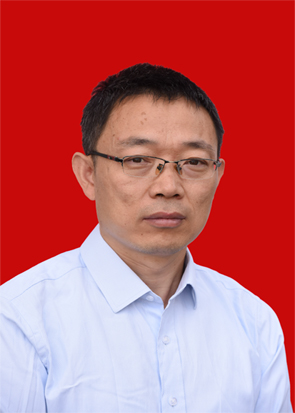}}]
	{Fuyuan Cao}  received the M.S. and Ph.D. degrees in computer science from Shanxi University, Taiyuan, China, in 2004 and 2010, respectively. He is currently a professor with the school of computer and information technology, Shanxi University, China. 
	
	His current research interests include machine learning  and clustering analysis.
\end{IEEEbiography}
\vspace{-30pt}
\begin{IEEEbiography}[{\includegraphics[width=1in,height=1.2in,clip,keepaspectratio]{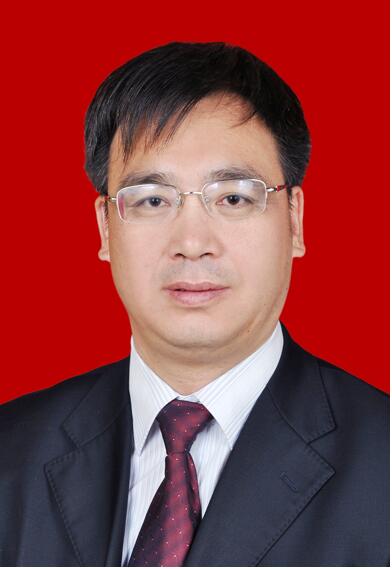}}]
	{Jiye Liang} (Senior Member, IEEE) received the M.S. and Ph.D. degrees from Xi’an Jiaotong University, Xi’an, China, in 1990 and 2001, respectively. He is currently a professor with the School of Computer and Information Technology, Shanxi University, Taiyuan, China, where he is also the director of the Key Laboratory of Computational Intelligence and Chinese Information Processing of the Ministry of Education. 
	
	He has authored more than 170 journal papers in his research fields. His current research interests include computational intelligence, granular computing, data mining, and knowledge discovery.
\end{IEEEbiography}

\clearpage
\begin{appendices}
\textbf{Supplementary Material} for ``Towards Privacy-Aware Causal Structure Learning in Federated Setting"
\section*{S-1: Detailed experimental results.}\label{secS-1} 
\subsection*{S-1-1: Experiment settings.}\label{secS-1-1} 
We implement the typical federated setting where each client owns its local data and don't communicate with other clients. Then clients transmits/receives only model parameter to/from the central server.
\subsubsection*{Datasets.}
The datasets used in the experiments include synthetic and real datasets as follows.

We assume that there are $K$ data samples in a dataset and $N$ clients exists, and $N$ lies in $\{3, 5, 10, 15\}$. To introduce unevenness, we randomly assigned the data samples to each client while ensuring that each client contains at least $\frac{K}{{N*2}}$ data samples. This approach aims to ensure that the data distribution is not heavily skewed towards any specific client.

\begin{itemize}
	\item \textbf{Synthetic datasets.} 
	We conduct FedPC on linear and nonlinear datasets. For the linear synthetic datasets, we generated five continuous datasets using an open-source toolkit \cite{kalainathan2020causal}  with the number of variables to 10, 20, 50, 100 and 300 respectively. Each synthetic dataset contains 5000 continuous samples. The generative process employed a linear causal mechanism represented as follows:
	\begin{equation}
		y = \textbf{X}W+\times E,
		\label{eq-0}
	\end{equation}
	where $+\times$ denotes either addition or multiplication, \textbf{X} denotes the vector of causes, and $E$ represents the noise variable accounting for all unobserved variables.
	For the nonlinear synthetic datasets, the causal mechanism used in the generative process is Gaussian Process (GP), and the mechanisms are represented as: 
	\begin{equation}
		y = GP(\textbf{X})+\times E.
		\label{eq-0}
	\end{equation}
	The proportion of noise in the mechanisms is set to 0.4. Gaussian noise was used in the generative process. In our experiments on nonlinear datasets, we utilized the Kernel-based Conditional Independence test (KCI-test) \cite{zhang2012kernel} instead of Fisher's Z Conditional Independence test. This was done to achieve better performance in detecting nonlinear dependencies in the data.
	\item \textbf{Benchmark Bayesian network (BN) datasets.}
	We use five benchmark BNs, alarm, insurance, win95pts, andes and pigs, to generate five discrete datasets, respectively. Each dataset contained 5000 samples. The details of the five benchmark BNs are presented in Table 1.
	\item \textbf{Real dataset.} We also compare FedPC with its rivals on a real biological dataset with 853 samples, Sachs~\cite{sachs2005causal}. Sachs is a protein signaling network expressing the level of different proteins and phospholipids in human cells. It is commonly viewed as a benchmark graphical model with 11 nodes (cell types) and 17 edges.
	Same as above, data samples are distributed unevenly across $N$ clients ($N$ $\in$ \{3, 5, 10, 15\}), and the number of samples on each client is randomly generated and is at least $\frac{N}{{K*2}}$.
\end{itemize}

\begin{table}
	\caption{Details of five benchmark Bayesian networks.}
	\centering
	\begin{tabular}{c c c c}
		\toprule
		network & \tabincell{c}{Number of\\variables} & \tabincell{c}{Number of\\edges} & \tabincell{c}{Maximum\\in/out-degree} \\
		\midrule
		alarm & 37 & 46 & 4/5 \\
		insurance & 27 & 52 & 3/7 \\
		win95pts & 76 & 112 & 7/10 \\
		andes & 223 & 338 & 6/12 \\
		pigs & 441 & 592 & 2/39 \\
		\bottomrule
	\end{tabular}
\end{table}

\subsubsection*{Metrics.}
To evaluate the performance of FedPC, we use the following frequently used metrics in DAG learning.
\begin{itemize}
	\item Reverse: Reverse is the number of edges with wrong directions in the DAG learnt by an algorithm against the true DAG.
	\item Extra: Extra is the number of extra edges in the DAG learnt by an algorithm against the true DAG.
	\item Miss: Miss is the number of missing edges in the DAG learnt by an algorithm against the true DAG.
	\item SHD (Structural Hamming Distance): The value of SHD is calculated by comparing the learnt causal structure with the true causal structure. Specifically, the value of SHD is the sum of undirected edges, reverse edges, missing edges and extra edges.
	\item TPR: TPR is also called recall, which refers to the probability that an actual positive will test positive. TPR is calculated as: $TPR = \frac{{TP}}{{TP + FN}}$
	\item FDR: FDR is the expected ratio of the number of false positive classifications to the total number of positive classifications: $FDR = \frac{{FP}}{{TP + FP}}$
	\item Precision: Precision denotes the number of true positives in the output divided by the number of features in the output of the algorithm. Precision is calculated as: $Precision = \frac{{TP}}{{TP + FP}}$
\end{itemize}

FedPC utilizes the PC algorithm to learn the skeleton and orient the undirected edges in the causal graph. Therefore, the output of FedPC is a CPDAG, which contains both directed and undirected edges. We employ the acyclic constraint technology proposed in reference\cite{spirtes2000causation} to transform CPDAGs returned by FedPC into DAGs, and then calculated reverse, extra, miss, shd, TPR, FDR and precision scores of these DAGs. In Tables~\ref{tbl-ala}-\ref{tbl-sachs}, the symbol ``-" denotes that an algorithm does not produce results on the corresponding datasets when the running time of the algorithm exceeded 72 hours or there is no enough memory space.

\subsubsection*{Baselines and rivals.}
FedPC is compared with 8 rivals. The comparison methods are as follows:

(1) NOTEARS-Avg. We run the NOTEARS~\cite{zheng2018dags} algorithm at each client independently, then calculate the averaging results of SHD and F1 of all learnt DAGs as the final results. 

(2) NOTEARS-ADMM. We run the NOTEARS-ADMM algorithm~\cite{ng2022towards}, and then calculate the SHD and F1 of the learnt DAG as the final results.

(3) FedDAG. We run the FedDAG~\cite{gao2023feddag} algorithm and then calculate the SHD and F1 of the learnt DAG as the final results.

(4) PC-Avg. We first run the PC algorithm at each client independently for obtaining $N$ DAGs ($N$ is the number of clients), and then calculate the averaging SHD values and F1 values of all learnt DAGs as the final results of PC-Avg. 

(5) PC-Best. We first run the PC algorithm at each client independently to get $N$ DAGs, and then select the DAG with the lowest SHD value as the final output.

(6) PC-All. We centralize all clients’ data to a single dataset and run the PC algorithm on it.

(7) FedPC-Simple-I. We run the PC algorithm at each client independently to learn the DAGs, then aggregate all learnt DAGs at the server by the strategy that if more than 30\% (the same ratio as our method) of the learnt DAGs contain a directed edge between two variables, this edge is kept in the final DAG.

(8) FedPC-Simple-II. We run the PC algorithm to learn the skeletons independently at each client, then aggregate all learnt skeletons at the server by the strategy that if an undirected edge between two variables exists on more than 30\% of the skeletons, this edge will be kept in the final skeleton.
Then we take the intersection of the separation sets between two variables learnt from each client to learn v-structures.
This is an ablation study of our proposed algorithm, by removing the layer-wise strategy of the FedOrien subroutine.
\subsubsection*{Implementation details.}
All experiments were conducted on a computer with Intel Core i9-10900F 2.80-GHz CPU and 32-GB memory. The significance level for CI tests is set to 0.01. For PC, NOTEARS, and NOTEARS-ADMM, we used the source codes provided by their authors. NOTEARS-Avg and NOTEARS-ADMM use 0.3 as the threshold to prune edges in a DAG and FedDAG uses 0.5 as the threshold, those are the same as the original paper. 

\subsection*{S-1-2: Experiment results on benchmark data.}\label{secS-1-2} 
\subsubsection*{Structural errors.}
Table \ref{tbl-ala} to \ref{tbl-300n} show the reverse, extra, miss, SHD values of FedPC and its rivals using five benchmark BN datasets and five continuous synthetic datasets, respectively.

\begin{itemize}
	\item Generally, we can see that FedPC achieves a lower SHD than its rivals, which indicates the superiority of our method. The reason is as follows: The excellent performance of FedPC relies on correctly learnt skeleton structures, because the layer-wise aggregation strategy enables FedPC to reduce the number of miss edges. And simultaneously the consistent separation set identification strategy further lessens the number of reverse edges.
	
	\item For all datasets, NOTEARS-Avg and NOTEARS-ADMM have similar performance. But we notice that NOTEARS-Avg and NOTEARS-ADMM perform good at reverse and extra edges on almost all datasets. The explanation is that the size of the learnt DAGs is much smaller than others, leading that NOTEARS-Avg and NOTEARS-ADMM miss many true edges (i.e. having higher miss values and higher SHD values). So the final DAGs exist a small amount of reverse and extra edges.
	
	\item It can be seen that the quality of the DAGs learnt by FedPC-Simple-I is competitive with that learnt by and FedPC-Simple-II. Inaccurate orientation leads that FedPC-Simple-I and FedPC-Simple-II are inferior in reverse. FedPC-Simple-I orients edges by simply using a voting scheme to determine the orientation. FedPC-Simple-II directly employs the intersection-rule to construct the separation set. As a result, FedPC-Simple-II avoids learning many correctly oriented edges in the DAG.
	
	\item The DAGs learnt by PC-Avg and PC-Best have more extra and reverse edges than those learnt by FedPC, which indicates that FedPC finds more accurate skeletons than PC-Avg and PC-Best. An explanation might be that PC-Avg and PC-Best do not exchange information between clients while FedPC exchanges its learnt skeleton with other clients at each layer. This verifies the effectiveness of FedPC and indicates that exchanging information during the skeleton learning process is a key to learning an accurate DAG, as is the case for FedPC.
	
\end{itemize}

\subsubsection*{Structural correctness.}
Through the metrics of structural correctness, i.e. TPR, FDR and precision, Table \ref{tbl-ala} to \ref{tbl-300n} report the quality of DAG learnt by different algorithms on five benchmark BN datasets and five continuous synthetic datasets, respectively.
We find that on most datasets, FedPC not only achieves fewer structural errors than its rivals, bust also achieves more structural correctness than other algorithms on almost all datasets.

\begin{itemize}
	\item Compared with its rivals, FedPC obtains highest values of TPR on most benchmark datasets and most synthetic datasets. Specifically, for TPR metric, our method achieves clear improvements of approximately 58\% more than NOTEARS-ADMM on alarm with 5000 samples when the number of clients is 3, 50\% more than NOTEARS-Avg on insurance with 5000 samples when the number of clients is 10, 10\% more than FedPC-Simple-I, PC-Best, PC-Avg on win95pts with 5000 samples when the number of clients is 15. The reason is that FedPC adopts the layer-wise skeleton learning strategy to construct more accurate skeleton, that is, some missed edges are restored and some extra edges are removed.
	\item For precision metric, our method achieves clear improvements on high-dimensional data, the explanation may be that FedPC gets more correct edges in the learnt DAG. And the novel method for separation sets employed by our method, which can recover some missed true directed edges, thus FedPC achieves a higher precision value.
	\item FedPC obtains lowest values of FDR on most benchmark datasets and most synthetic datasets. FDR measures the number of false positive items indicating that there exiting an edge in the predicted DAG but not in the true DAG. FedPC, employing a novel strategy to find out the true separation set, removes some redundant false directed edges.
	
	\item PC-Avg and PC-Best have little improvement on TPR, FDR and precision can be explained that PC-Avg and PC-Best do not exchange information between clients, they learns more wrong edges because of insufficient data samples. But the layer-wise aggregation strategy makes FedPC exchange enough information between clients for accurate DAG learning. This further verifies the effectiveness of the two strategies of FedPC.
	
	\item In addition, we find that the performance of continuous optimization methods (such as NOTEARS-Avg and NOTEARS-ADMM) is generally worse than that of combinatorial optimization methods (such as PC-Avg, PC-Best, FedPC-Simple-I, FedPC-Simple-II) on most benchmark datasets in terms of TPR, FDR and precision. This is because that the causal structures learnt by continuous optimization methods contain many extra edges and reverse edges. More specifically, the TPR is greatly reduced and the FDR is significantly promoted.
	
	\item FedPC-Simple-I and FedPC-Simple-II performs worse compared with PC-Avg and PC-Best. In particular, FedPC-Simple-I and FedPC-Simple-II is not suitable to datasets with large in/out-degrees,such as Pigs. The explanation is that FedPC-Simple-I and FedPC-Simple-II lack a efficient way to identify the correct separation set when orienting edges. And because of the decreased scale of data samples, CI tests become inaccurate, leading the performance of PC-Avg and PC-Best .
	
\end{itemize}

\subsection*{S-1-3: Experiment results on nonlinear data.}\label{secS-1-3} 

In order to evaluate the performance of FedPC on nonlinear datasets, we conducted additional experiments using synthetic data generated with nonlinear causal mechanism. Specifically, we used Kernel-based Conditional Independence (KCI) tests \cite{zhang2012kernel} instead of Fisher's Z Conditional Independence test to achieve ideal performance in the nonlinear setting. The KCI tests have been shown to be effective in detecting nonlinear causal relationships in previous studies.

We evaluated the performance of FedPC on the nonlinear datasets using metrics such as Structural Hamming Distance (SHD), True Positive Rate (TPR), precision, and F1 score. Due to the slow running time of the KCI test, we conducted experiments on causal graphs with 10 and 20 nodes only. To provide a comprehensive comparison, we also included FedDAG and NOTEARS-ADMM in the comparison. Both FedDAG and NOTEARS-ADMM have been shown to support nonlinear relationships \cite{ng2022towards}\cite{gao2023feddag}, making them suitable for comparison with FedPC in the nonlinear setting.

The results of the experiments are shown in Table \ref{tbl-nlnr}. As can be seen from the table, FedPC achieved the lowest SHD and highest F1 score among all three algorithms, demonstrating its effectiveness in handling nonlinear datasets. 

The experiments show that FedPC is not only effective in handling linear datasets, but also performs well on nonlinear datasets, making it a versatile algorithm for federated learning.

\begin{table*}[htbp]
	\caption{Details of benchmark Bayesian network: alarm}
	\label{tbl-ala}
	\centering
	\scriptsize
	\setlength{\tabcolsep}{2mm}{
		\begin{tabular}{ccccccccc}
			\toprule 
			\tabincell{c}{Number of\\clients}   & Algorithm    & Reverse($\downarrow$) & Extra($\downarrow$)  & Miss($\downarrow$)  & SHD($\downarrow$)    & TPR ($\uparrow$)   & FDR($\downarrow$)   & Precision($\uparrow$)\\
			\midrule 
			\multirow{7}{*}{3}  & PC-Avg          & 9       & 9.667  & 4     & 22.667 & 0.717 & 0.361 & 0.639     \\
			& PC-Best         & 9       & 9      & \textbf{3}     & 21     & 0.739 & 0.346 & 0.654     \\
			& FedPC-Simple-I  & 5       & \textbf{0}      & 4     & 9      & 0.804 & 0.119 & 0.881     \\
			& FedPC-Simple-II & 8       & \textbf{0}      & 4     & 12     & 0.739 & 0.190 & 0.810     \\
			& NOTEARS-Avg     & \textbf{3}       & 4      & 36    & 43     & 0.152 & 0.500 & 0.500     \\
			& NOTEARS-ADMM    & 5       & 17     & 28    & 50     & 0.283 & 0.629 & 0.371     \\		 
			& {FedDAG} & {46}                             & {46}                           & {0}                           & {92}                         & {\textbf{1.000}}                           & {0.667}                     & {0.333}                           \\
			& FedPC          & \textbf{3}       & \textbf{0}      & \textbf{3}     & \textbf{6}      & 0.870 & \textbf{0.070} & \textbf{0.930}     \\
			\midrule 
			\multirow{7}{*}{5}  & PC-Avg          & 9.4     & 13.8   & 7.2   & 30.4   & 0.639 & 0.430 & 0.570     \\
			& PC-Best         & 7       & 10     & \textbf{6}     & 23     & 0.717 & 0.340 & 0.660     \\
			& FedPC-Simple-I  & 7       & \textbf{4}      & \textbf{6}     & \textbf{17}     & 0.717 & \textbf{0.250} & \textbf{0.750}     \\
			& FedPC-Simple-II & 14      & \textbf{4}      & \textbf{6}     & 24     & 0.565 & 0.409 & 0.591     \\
			& NOTEARS-Avg     & \textbf{3}       & \textbf{4}      & 37    & 44     & 0.130 & 0.538 & 0.462     \\
			& NOTEARS-ADMM    & 4       & 13     & 27    & 44     & 0.326 & 0.531 & 0.469     \\
			& {FedDAG} & {46}                             & {46}                           & {0}                           & {92}                         & \textbf{{1.000}}                           & {0.667}                     & {0.333}                           \\
			& FedPC           & 7       & 7      & \textbf{6}     & 20     & 0.717 & 0.298 & 0.702     \\
			\midrule 
			\multirow{7}{*}{10} & PC-Avg          & 10.3    & 26.3   & 8.2   & 44.8   & 0.598 & 0.535 & 0.465     \\
			& PC-Best         & 10      & 9      & 7     & 26     & 0.630 & 0.396 & 0.604     \\
			& FedPC-Simple-I  & 11      & \textbf{7}      & 9     & 27     & 0.565 & 0.409 & 0.591     \\
			& FedPC-Simple-II & 15      & \textbf{7}      & 9     & 31     & 0.478 & 0.500 & 0.500     \\
			& NOTEARS-Avg     & 5       & 5      & 36    & 46     & 0.109 & 0.667 & 0.333     \\
			& NOTEARS-ADMM    & \textbf{4}       & 10     & 28    & 42     & 0.304 & 0.500 & 0.500     \\
			& {FedDAG} & {46}                             & {46}                           & {0}                           & {92}                         & \textbf{{1.000}}                           & {0.667}                     & {0.333}                           \\
			& FedPC           & 10      & \textbf{7}      & \textbf{6}     & \textbf{23}     & 0.652 & \textbf{0.362} & \textbf{0.638}     \\
			\midrule 
			\multirow{7}{*}{15} & PC-Avg          & 11.2    & 49.133 & 9.933 & 70.267 & 0.541 & 0.685 & 0.315     \\
			& PC-Best         & 10      & 24     & \textbf{7}     & 41     & 0.630 & 0.540 & 0.460     \\
			& FedPC-Simple-I  & 10      & 7      & 9     & 26     & 0.587 & 0.386 & 0.614     \\
			& FedPC-Simple-II & 13      & 7      & 9     & 29     & 0.522 & 0.455 & 0.545     \\
			& NOTEARS-Avg     & 5       & \textbf{4}      & 36    & 45     & 0.109 & 0.643 & 0.357     \\
			& NOTEARS-ADMM    & \textbf{4}       & 12     & 29    & 45     & 0.283 & 0.552 & 0.448     \\
			& {FedDAG} & {46}                             & {46}                           & {0}                           & {92}                         & \textbf{{1.000}}                           & {0.667}                     & {0.333}                           \\
			& FedPC           & 6       & 7      & 9     & \textbf{22}     & 0.674 & \textbf{0.295} & \textbf{0.705}	  \\
			\bottomrule 
	\end{tabular}}
\end{table*}

\begin{table*}[htbp]
	\caption{Details of benchmark Bayesian network: insurance}
	\label{tbl-ins}
	\centering
	\scriptsize
	\setlength{\tabcolsep}{2mm}{
		\begin{tabular}{ccccccccc}
			\toprule 
			Number of clients   & Algorithm    & Reverse($\downarrow$) & Extra($\downarrow$)  & Miss($\downarrow$)  & SHD($\downarrow$)    & TPR ($\uparrow$)   & FDR($\downarrow$)   & Precision($\uparrow$)\\
			\midrule 
			\multirow{7}{*}{3}  & PC-Avg          & 7.333   & 6.333  & 17.667 & 31.333 & 0.519 & 0.335 & 0.665     \\
			& PC-Best         & 9       & 4      & \textbf{16}     & 29     & 0.519 & 0.325 & 0.675     \\
			& FedPC-Simple-I  & 6       & \textbf{0}      & 19     & 25     & 0.519 & 0.182 & 0.818     \\
			& FedPC-Simple-II & 8       & \textbf{0}      & 19     & 27     & 0.481 & 0.242 & 0.758     \\
			& NOTEARS-Avg     & \textbf{3}       & 2      & 44     & 49     & 0.096 & 0.500 & 0.500     \\
			& NOTEARS-ADMM    & 6       & 31     & 35     & 72     & 0.212 & 0.771 & 0.229     \\
			& {FedDAG} & {52}                             & {52}                           & {0}                           & {104}                        & \textbf{{1.000}}                           & {0.667}                     & {0.333}                           \\
			& FedPC           & 6       & \textbf{0}      & 18     & \textbf{24}     & 0.538 & \textbf{0.176} & \textbf{0.824}     \\
			\midrule 
			\multirow{7}{*}{5}  & PC-Avg          & 7.6     & 12.8   & 19.2   & 39.6   & 0.485 & 0.445 & 0.555     \\
			& PC-Best         & 6       & 10     & \textbf{19}     & 35     & 0.519 & 0.372 & 0.628     \\
			& FedPC-Simple-I  & 5       & 2      & 21     & 28     & 0.500 & 0.212 & 0.788     \\
			& FedPC-Simple-II & 11      & \textbf{1}      & 21     & 33     & 0.385 & 0.375 & 0.625     \\
			& NOTEARS-Avg     & \textbf{3}       & 2      & 44     & 49     & 0.096 & 0.500 & 0.500     \\
			& NOTEARS-ADMM    & 5       & 27     & 35     & 67     & 0.231 & 0.727 & 0.273     \\
			& {FedDAG} & {52}                             & {52}                           & {0}                           & {104}                        & \textbf{{1.000}}                           & {0.667}                     & {0.333}                           \\
			& FedPC           & \textbf{3}       & 2      & 21     & \textbf{26}     & 0.538 & \textbf{0.152} & \textbf{0.848}     \\
			\midrule 
			\multirow{7}{*}{10} & PC-Avg          & 10.5    & 26.3   & 18.9   & 55.7   & 0.435 & 0.594 & 0.406     \\
			& PC-Best         & 6       & 11     & \textbf{19}     & 36     & 0.519 & 0.386 & 0.614     \\
			& FedPC-Simple-I  & 8       & 8      & 22     & 38     & 0.423 & 0.421 & 0.579     \\
			& FedPC-Simple-II & 9       & 7      & 22     & 38     & 0.404 & 0.432 & 0.568     \\
			& NOTEARS-Avg     & \textbf{3}       & \textbf{2}      & 44     & 49     & 0.096 & 0.500 & 0.500     \\
			& NOTEARS-ADMM    & 5       & 20     & 39     & 64     & 0.154 & 0.758 & 0.242     \\
			& {FedDAG} & {52}                             & {52}                           & {0}                           & {104}                        & \textbf{{1.000}}                           & {0.667}                     & {0.333}                           \\
			& FedPC           & 4       & 7      & 22     & \textbf{33}     & 0.500 & \textbf{0.297} & \textbf{0.703}     \\
			\midrule 
			\multirow{7}{*}{15} & PC-Avg          & 12.867  & 47.133 & 17.867 & 77.867 & 0.409 & 0.728 & 0.272     \\
			& PC-Best         & 12      & 19     & \textbf{19}     & 50     & 0.404 & 0.596 & 0.404     \\
			& FedPC-Simple-I  & 11      & 13     & 21     & 45     & 0.385 & 0.545 & 0.455     \\
			& FedPC-Simple-II & 11      & 8      & 23     & 42     & 0.346 & 0.514 & 0.486     \\
			& NOTEARS-Avg     & \textbf{3}       & \textbf{1}      & 44     & 48     & 0.096 & 0.444 & 0.556     \\
			& NOTEARS-ADMM    & 5       & 16     & 39     & 60     & 0.154 & 0.724 & 0.276     \\
			& {FedDAG} & {52}                             & {52}                           & {0}                           & {104}                        & \textbf{{1.000}}                           & {0.667}                     & {0.333}                           \\
			& FedPC           & 7       & 9      & 22     & \textbf{38}     & 0.442 & \textbf{0.410} & \textbf{0.590}	   \\
			\bottomrule 
	\end{tabular}}
\end{table*}

\begin{table*}[htbp]
	\caption{Details of benchmark Bayesian network: win95pts}
	\label{tbl-win}
	\centering
	\scriptsize
	\setlength{\tabcolsep}{2mm}{
		\begin{tabular}{ccccccccc}
			\toprule 
			Number of clients   & Algorithm    & Reverse($\downarrow$) & Extra($\downarrow$)  & Miss($\downarrow$)  & SHD($\downarrow$)    & TPR ($\uparrow$)   & FDR($\downarrow$)   & Precision($\uparrow$)\\
			\midrule 
			\multirow{7}{*}{3}  & PC-Avg          & 4.333   & 2.333 & 59.667 & 66.333 & 0.429 & 0.123 & 0.877     \\
			& PC-Best         & 5       & 1     & 54     & 60     & 0.473 & 0.102 & 0.898     \\
			& FedPC-Simple-I  & 5       & 6     & \textbf{46}     & 57     & 0.545 & 0.153 & 0.847     \\
			& FedPC-Simple-II & 5       & 6     & \textbf{46}     & 57     & 0.545 & 0.153 & 0.847     \\
			& NOTEARS-Avg     & \textbf{2}       & \textbf{0}     & 110    & 112    & 0.000 & 1.000 & 0.000     \\
			& NOTEARS-ADMM    & 4       & 5     & 94     & 103    & 0.125 & 0.391 & 0.609     \\
			& {FedDAG} & {-}                              & {-}                            & {-}                           & {-}                          & {-}                           & {-}                          & {-}                                \\
			& FedPC           & \textbf{2}       & 5     & \textbf{46}     & \textbf{53}     & \textbf{0.571} & \textbf{0.099} & \textbf{0.901}     \\
			\midrule 
			\multirow{7}{*}{5}  & PC-Avg          & 5.2     & 3.4   & 67.2   & 75.8   & 0.354 & 0.182 & 0.818     \\
			& PC-Best         & 5       & 2     & \textbf{59}     & \textbf{66}     & \textbf{0.429} & 0.127 & 0.873     \\
			& FedPC-Simple-I  & 2       & 3     & 62     & 67     & \textbf{0.429} & 0.094 & 0.906     \\
			& FedPC-Simple-II & 4       & 3     & 62     & 69     & 0.411 & 0.132 & 0.868     \\
			& NOTEARS-Avg     & \textbf{0}       & \textbf{0}     & 111    & 111    & 0.009 & \textbf{0.000} & \textbf{1.000}     \\
			& NOTEARS-ADMM    & \textbf{0}       & 3     & 100    & 103    & 0.107 & 0.200 & 0.800     \\
			& {FedDAG} & {-}                              & {-}                            & {-}                           & {-}                          & {-}                           & {-}                          & {-}                                \\
			& FedPC           & 6       & 4     & 62     & 72     & 0.393 & 0.185 & 0.815     \\
			\midrule 
			\multirow{7}{*}{10} & PC-Avg          & 3.5     & 4.5   & 74.6   & 82.6   & 0.303 & 0.190 & 0.810     \\
			& PC-Best         & 4       & 6     & \textbf{65}     & 75     & 0.384 & 0.189 & 0.811     \\
			& FedPC-Simple-I  & 3       & 4     & 66     & 73     & 0.384 & 0.140 & 0.860     \\
			& FedPC-Simple-II & 7       & 4     & 66     & 77     & 0.348 & 0.220 & 0.780     \\
			& NOTEARS-Avg     & \textbf{1}       & \textbf{0}     & 111    & 112    & 0.000 & 1.000 & 0.000     \\
			& NOTEARS-ADMM    & \textbf{1}       & 2     & 102    & 105    & 0.080 & 0.250 & 0.750     \\
			& {FedDAG} & {-}                              & {-}                            & {-}                           & {-}                          & {-}                           & {-}                          & {-}                                \\
			& FedPC           & 3       & 4     & \textbf{65}     & \textbf{72}     & \textbf{0.393} & \textbf{0.137} & \textbf{0.863}     \\
			\midrule 
			\multirow{7}{*}{15} & PC-Avg          & 3.067   & 5.133 & 80     & 88.2   & 0.258 & 0.218 & 0.782     \\
			& PC-Best         & 2       & 3     & 75     & 80     & 0.313 & \textbf{0.125} & \textbf{0.875}     \\
			& FedPC-Simple-I  & 3       & 3     & 74     & 80     & 0.313 & 0.146 & 0.854     \\
			& FedPC-Simple-II & 8       & 3     & 74     & 85     & 0.268 & 0.268 & 0.732     \\
			& NOTEARS-Avg     & \textbf{1}       & \textbf{0}     & 111    & 112    & 0.000 & 1.000 & 0.000     \\
			& NOTEARS-ADMM    & \textbf{1}       & 2     & 105    & 108    & 0.054 & 0.333 & 0.667     \\
			& {FedDAG} & {-}                              & {-}                            & {-}                           & {-}                          & {-}                           & {-}                          & {-}                                \\
			& FedPC           & 4       & 3     & \textbf{63}     & \textbf{70}     & \textbf{0.402} & 0.135 & 0.865     \\
			\bottomrule 
	\end{tabular}}
\end{table*}

\begin{table*}[htbp]
	\caption{Details of benchmark Bayesian network: andes}
	\label{tbl-and}
	\centering
	\scriptsize
	\setlength{\tabcolsep}{2mm}{
		\begin{tabular}{ccccccccc}
			\toprule 
			Number of clients   & Algorithm    & Reverse($\downarrow$) & Extra($\downarrow$)  & Miss($\downarrow$)  & SHD($\downarrow$)    & TPR ($\uparrow$)   & FDR($\downarrow$)   & Precision($\uparrow$)\\
			\midrule 
			\multirow{7}{*}{3}  & PC-Avg          & 2.333   & 10     & 115.333 & 127.667 & 0.652 & 0.053 & 0.947     \\
			& PC-Best         & \textbf{1}       & 7      & 117     & 125     & 0.651 & \textbf{0.035} & \textbf{0.965}     \\
			& FedPC-Simple-I  & 3       & 28     & 99      & 130     & 0.698 & 0.116 & 0.884     \\
			& FedPC-Simple-II & 2       & 28     & 99      & 129     & 0.701 & 0.112 & 0.888     \\
			& NOTEARS-Avg     & 3       & \textbf{0}      & 325     & 328     & 0.030 & 0.231 & 0.769     \\
			& NOTEARS-ADMM    & 30      & 2      & 284     & 316     & 0.071 & 0.571 & 0.429     \\
			& {FedDAG} & {-}                              & {-}                            & {-}                           & {-}                          & {-}                           & {-}                          & {-}                                \\
			& FedPC           & 2       & 23     & \textbf{98}      & \textbf{123}     & \textbf{0.704} & 0.095 & 0.905     \\
			\midrule 
			\multirow{7}{*}{5}  & PC-Avg          & 1.8     & 10.8   & 127.2   & 139.8   & 0.618 & 0.057 & 0.943     \\
			& PC-Best         & 1       & 8      & 117     & 126     & \textbf{0.651} & 0.039 & 0.961     \\
			& FedPC-Simple-I  & \textbf{0}       & 1      & 119     & \textbf{120}     & 0.648 & \textbf{0.005} & \textbf{0.995}     \\
			& FedPC-Simple-II & 1       & 1      & 119     & 121     & 0.645 & 0.009 & 0.991     \\
			& NOTEARS-Avg     & 3       & \textbf{0}      & 326     & 329     & 0.027 & 0.250 & 0.750     \\
			& NOTEARS-ADMM    & 23      & 2      & 292     & 317     & 0.068 & 0.521 & 0.479     \\
			& {FedDAG} & {-}                              & {-}                            & {-}                           & {-}                          & {-}                           & {-}                          & {-}                                \\
			& FedPC           & 4       & 1      & \textbf{115}     & \textbf{120}     & 0.648 & 0.022 & 0.978     \\
			\midrule 
			\multirow{7}{*}{10} & PC-Avg          & 4.8     & 22.2   & 152.3   & 179.3   & 0.535 & 0.128 & 0.872     \\
			& PC-Best         & 3       & 17     & 140     & 160     & 0.577 & 0.093 & 0.907     \\
			& FedPC-Simple-I  & \textbf{1}       & 2      & 142     & 145     & 0.577 & \textbf{0.015} & \textbf{0.985}     \\
			& FedPC-Simple-II & 4       & 2      & 142     & 148     & 0.568 & 0.030 & 0.970     \\
			& NOTEARS-Avg     & 5       & \textbf{0}      & 327     & 332     & 0.018 & 0.455 & 0.545     \\
			& NOTEARS-ADMM    & 20      & 1      & 307     & 328     & 0.033 & 0.656 & 0.344     \\
			& {FedDAG} & {-}                              & {-}                            & {-}                           & {-}                          & {-}                           & {-}                          & {-}                                \\
			& FedPC           & 5       & 3      & \textbf{134}     & \textbf{142}     & \textbf{0.589} & 0.039 & 0.961     \\
			\midrule 
			\multirow{7}{*}{15} & PC-Avg          & 6.933   & 32.133 & 158.4   & 197.467 & 0.511 & 0.185 & 0.815     \\
			& PC-Best         & 3       & 29     & \textbf{147}     & 179     & \textbf{0.556} & 0.145 & 0.855     \\
			& FedPC-Simple-I  & \textbf{1}       & 3      & 166     & \textbf{170}     & 0.506 & \textbf{0.023} & \textbf{0.977}     \\
			& FedPC-Simple-II & \textbf{1}       & 3      & 166     & \textbf{170}     & 0.506 & \textbf{0.023} & \textbf{0.977}     \\
			& NOTEARS-Avg     & 4       & \textbf{0}      & 327     & 331     & 0.021 & 0.364 & 0.636     \\
			& NOTEARS-ADMM    & 13      & \textbf{0}      & 315     & 328     & 0.030 & 0.565 & 0.435     \\
			& {FedDAG} & {-}                              & {-}                            & {-}                           & {-}                          & {-}                           & {-}                          & {-}                                \\
			& FedPC           & 4       & 3      & 164     & 171     & 0.503 & 0.040 & 0.960     \\
			\bottomrule 
	\end{tabular}}
\end{table*}

\begin{table*}[htbp]
	\caption{Details of benchmark Bayesian network: pigs}
	\label{tbl-pig}
	\centering
	\scriptsize
	\setlength{\tabcolsep}{2mm}{
		\begin{tabular}{ccccccccc}
			\toprule 
			Number of clients   & Algorithm    & Reverse($\downarrow$) & Extra($\downarrow$)  & Miss($\downarrow$)  & SHD($\downarrow$)    & TPR ($\uparrow$)   & FDR($\downarrow$)   & Precision($\uparrow$)\\
			\midrule 
			\multirow{7}{*}{3}  & PC-Avg          & 59      & 6.333   & \textbf{0}    & 65.333  & 0.900 & 0.109 & 0.891     \\
			& PC-Best         & 1       & 8       & \textbf{0}    & \textbf{9}       & 0.998 & \textbf{0.015} & \textbf{0.985}     \\
			& FedPC-Simple-I  & 50      & 13      & \textbf{0}    & 63      & 0.916 & 0.104 & 0.896     \\
			& FedPC-Simple-II & 127     & 13      & \textbf{0}    & 140     & 0.785 & 0.231 & 0.769     \\
			& NOTEARS-Avg     & 59      & \textbf{0}       & 379  & 438     & 0.260 & 0.277 & 0.723     \\
			& NOTEARS-ADMM    & 180     & 22      & 60   & 262     & 0.595 & 0.365 & 0.635     \\
			& {FedDAG} & {-}                              & {-}                            & {-}                           & {-}                          & {-}                           & {-}                          & {-}                                \\
			& FedPC           & \textbf{0}       & 10      & \textbf{0}    & 10      & \textbf{1.000} & 0.017 & 0.983     \\
			\midrule 
			\multirow{7}{*}{5}  & PC-Avg          & 193.6   & 821     & \textbf{0}    & 1014.6  & 0.673 & 0.499 & 0.501     \\
			& PC-Best         & \textbf{3}       & 4       & \textbf{0}    & \textbf{7}       & \textbf{0.995} & \textbf{0.012} & \textbf{0.988}     \\
			& FedPC-Simple-I  & 162     & 3       & \textbf{0}    & 165     & 0.726 & 0.277 & 0.723     \\
			& FedPC-Simple-II & 342     & 3       & \textbf{0}    & 345     & 0.422 & 0.580 & 0.420     \\
			& NOTEARS-Avg     & 24      & \textbf{0}       & 491  & 515     & 0.130 & 0.238 & 0.762     \\
			& NOTEARS-ADMM    & 181     & 19      & 72   & 272     & 0.573 & 0.371 & 0.629     \\
			& {FedDAG} & {-}                              & {-}                            & {-}                           & {-}                          & {-}                           & {-}                          & {-}                                \\
			& FedPC           & 16      & 6       & \textbf{0}    & 22      & 0.973 & 0.037 & 0.963     \\
			\midrule 
			\multirow{7}{*}{10} & PC-Avg          & 257.8   & 584.5   & \textbf{0}    & 842.3   & 0.565 & 0.594 & 0.406     \\
			& PC-Best         & 229     & 175     & \textbf{0}    & 404     & 0.613 & 0.527 & 0.473     \\
			& FedPC-Simple-I  & 236     & 1       & \textbf{0}    & 237     & 0.601 & 0.400 & 0.600     \\
			& FedPC-Simple-II & 334     & 1       & \textbf{0}    & 335     & 0.436 & 0.565 & 0.435     \\
			& NOTEARS-Avg     & 30      & \textbf{0}       & 453  & 483     & 0.184 & 0.216 & 0.784     \\
			& NOTEARS-ADMM    & 174     & 13      & 161  & 348     & 0.434 & 0.421 & 0.579     \\
			& {FedDAG} & {-}                              & {-}                            & {-}                           & {-}                          & {-}                           & {-}                          & {-}                                \\
			& FedPC           & \textbf{7}       & 1       & \textbf{0}    & \textbf{8}       & \textbf{0.988} & \textbf{0.013} & \textbf{0.987}     \\
			\midrule 
			\multirow{7}{*}{15} & PC-Avg          & 304.933 & 1752.53 & \textbf{0}    & 2057.47 & 0.485 & 0.742 & 0.258     \\
			& PC-Best         & 213     & 200     & \textbf{0}    & 413     & 0.640 & 0.521 & 0.479     \\
			& FedPC-Simple-I  & 280     & 129     & \textbf{0}    & 409     & 0.527 & 0.567 & 0.433     \\
			& FedPC-Simple-II & 323     & 129     & \textbf{0}    & 452     & 0.454 & 0.627 & 0.373     \\
			& NOTEARS-Avg     & \textbf{23}      & \textbf{0}       & 484  & 507     & 0.144 & \textbf{0.213} & \textbf{0.787}     \\
			& NOTEARS-ADMM    & 143     & 2       & 238  & 383     & 0.356 & 0.407 & 0.593     \\
			& {FedDAG} & {-}                              & {-}                            & {-}                           & {-}                          & {-}                           & {-}                          & {-}                                \\
			& FedPC           & 120     & 135     & \textbf{0}    & \textbf{255}     & \textbf{0.797} & 0.351 & 0.649 	   \\
			\bottomrule 
	\end{tabular}}
\end{table*}

\begin{table*}[htbp]
	\caption{Details of continuous datasets: 10 nodes}
	\label{tbl-10n}
	\centering
	\scriptsize
	\setlength{\tabcolsep}{2mm}{
		\begin{tabular}{ccccccccc}
			\toprule 
			Number of clients   & Algorithm    & Reverse($\downarrow$) & Extra($\downarrow$)  & Miss($\downarrow$)  & SHD($\downarrow$)    & TPR ($\uparrow$)   & FDR($\downarrow$)   & Precision($\uparrow$)\\
			\midrule 
			\multirow{7}{*}{3}  & PC-Avg          & 2.333   & 0.667 & 3.667 & 6.667 & 0.600 & 0.246 & 0.754     \\
			& PC-Best         & \textbf{0}       & 1     & 5     & 6     & 0.667 & 0.091 & 0.909     \\
			& FedPC-Simple-I  & 1       & 1     & \textbf{3}     & \textbf{5}     & \textbf{0.733} & 0.154 & 0.846     \\
			& FedPC-Simple-II & 4       & 4     & \textbf{3}     & 11    & 0.533 & 0.500 & 0.500     \\
			& NOTEARS-Avg     & 3       & 1     & 9     & 13    & 0.200 & 0.571 & 0.429     \\
			& NOTEARS-ADMM    & 3       & 9     & 8     & 20    & 0.267 & 0.750 & 0.250     \\
			& {FedDAG} & {5}                              & {2}                            & {8}                           & {15}                         & {0.133}                    & {0.778}                   & {0.222}                         \\
			& FedPC           & \textbf{0}       & \textbf{0}     & 5     & \textbf{5}     & 0.667 & \textbf{0.000} & \textbf{1.000}     \\
			\midrule 
			\multirow{7}{*}{5}  & PC-Avg          & 1.8     & \textbf{0}     & 4.4   & 6.2   & 0.587 & 0.171 & 0.829     \\
			& PC-Best         & \textbf{0}       & \textbf{0}     & 4     & 4     & 0.733 & 0.000 & 1.000     \\
			& FedPC-Simple-I  & 1       & \textbf{0}     & 3     & 4     & 0.733 & 0.083 & 0.917     \\
			& FedPC-Simple-II & 6       & \textbf{0}     & \textbf{2}     & 8     & 0.467 & 0.462 & 0.538     \\
			& NOTEARS-Avg     & 2       & 1     & 11    & 14    & 0.133 & 0.600 & 0.400     \\
			& NOTEARS-ADMM    & 3       & 7     & 8     & 18    & 0.267 & 0.714 & 0.286     \\
			
			& {FedDAG} & {4}                              & {2}                            & {9}                           & {15}                         & {0.133}                      & {0.750}                     & {0.250}                             \\
			& FedPC           & \textbf{0}       & \textbf{0}     & 3     & \textbf{3}     & \textbf{0.800} & \textbf{0.000} & \textbf{1.000}     \\
			\midrule 
			\multirow{7}{*}{10} & PC-Avg          & 0.8     & 0.2   & 5.8   & 6.8   & 0.560 & 0.118 & 0.882     \\
			& PC-Best         & \textbf{0}       & \textbf{0}     & 4     & 4     & 0.733 & \textbf{0.000} & \textbf{1.000}     \\
			& FedPC-Simple-I  & \textbf{0}       & \textbf{0}     & 3     & \textbf{3}     & \textbf{0.800} & \textbf{0.000} & \textbf{1.000}     \\
			& FedPC-Simple-II & 6       & 5     & \textbf{2}     & 13    & 0.467 & 0.611 & 0.389     \\
			& NOTEARS-Avg     & 1       & \textbf{0}     & 12    & 13    & 0.133 & 0.333 & 0.667     \\
			& NOTEARS-ADMM    & 3       & 3     & 10    & 16    & 0.133 & 0.750 & 0.250     \\
			& {FedDAG} & {2}                              & {0}                            & {8}                           & {10}                         & {0.333}                      & {0.286}                     & {0.714}                           \\
			
			& FedPC           & \textbf{0}       & \textbf{0}     & 3     & \textbf{3}     & \textbf{0.800} & \textbf{0.000} & \textbf{1.000}     \\
			\midrule 
			\multirow{7}{*}{15} & PC-Avg          & 0.733   & 0.267 & 7.067 & 8.067 & 0.480 & 0.131 & 0.869     \\
			& PC-Best         & \textbf{0}       & \textbf{0}     & 4     & 4     & 0.733 & 0.000 & 1.000     \\
			& FedPC-Simple-I  & 1       & \textbf{0}     & 6     & 7     & 0.533 & 0.111 & 0.889     \\
			& FedPC-Simple-II & 5       & 2     & \textbf{1}     & 8     & 0.600 & 0.438 & 0.563     \\
			& NOTEARS-Avg     & 1       & \textbf{0}     & 12    & 13    & 0.133 & 0.333 & 0.667     \\
			& NOTEARS-ADMM    & 4       & 5     & 9     & 18    & 0.133 & 0.818 & 0.182     \\
			& {FedDAG} & {3}                              & {0}                            & {8}                           & {11}                         & {0.267}                      & {0.429}                     & {0.571}                           \\
			& FedPC           & \textbf{0}       & \textbf{0}     & 3     & \textbf{3}     & \textbf{0.800} & \textbf{0.000} & \textbf{1.000}     \\
			\bottomrule 
	\end{tabular}}
\end{table*}

\begin{table*}[htbp]
	\caption{Details of continuous datasets: 20 nodes}
	\label{tbl-20n}
	\centering
	\scriptsize
	\setlength{\tabcolsep}{2mm}{
		\begin{tabular}{ccccccccc}
			\toprule 
			Number of clients   & Algorithm    & Reverse($\downarrow$) & Extra($\downarrow$)  & Miss($\downarrow$)  & SHD($\downarrow$)    & TPR ($\uparrow$)   & FDR($\downarrow$)   & Precision($\uparrow$)\\
			\midrule 
			\multirow{7}{*}{3}  & PC-Avg          & 1.667   & 3.333 & 11.333 & 16.333 & 0.519 & 0.262 & 0.738     \\
			& PC-Best         & 2       & 3     & 10     & 15     & 0.556 & 0.250 & 0.750     \\
			& FedPC-Simple-I  & \textbf{1}       & 10    & 10     & 21     & \textbf{0.593} & 0.407 & 0.593     \\
			& FedPC-Simple-II & 4       & 19    & \textbf{9}      & 32     & 0.519 & 0.622 & 0.378     \\
			& NOTEARS-Avg     & 2       & 4     & 18     & 24     & 0.259 & 0.462 & 0.538     \\
			& NOTEARS-ADMM    & 6       & 35    & 16     & 57     & 0.185 & 0.891 & 0.109     \\
			& {FedDAG} & {27}                             & {29}                           & {0}                           & {56}                         & {1.000}                      & {0.675}                     & {0.325}                           \\
			
			& FedPC           & \textbf{1}       & \textbf{0}     & 13     & \textbf{14}     & 0.481 & \textbf{0.071} & \textbf{0.929}     \\
			\midrule 
			\multirow{7}{*}{5}  & PC-Avg          & 2.2     & 2.6   & 12.8   & 17.6   & 0.444 & 0.276 & 0.724     \\
			& PC-Best         & 1       & \textbf{0}     & 13     & 14     & 0.481 & 0.071 & 0.929     \\
			& FedPC-Simple-I  & 2       & 3     & 11     & 16     & 0.519 & 0.263 & 0.737     \\
			& FedPC-Simple-II & 4       & 5     & \textbf{8}      & 17     & \textbf{0.556} & 0.375 & 0.625     \\
			& NOTEARS-Avg     & \textbf{0}       & 3     & 25     & 28     & 0.074 & 0.600 & 0.400     \\
			& NOTEARS-ADMM    & 4       & 35    & 18     & 57     & 0.185 & 0.886 & 0.114     \\
			& {FedDAG} & {27}                             & {29}                           & {0}                           & {56}                         & {1.000}                      & {0.675}                     & {0.325}                           \\
			& FedPC           & 1       & \textbf{0}     & 11     & \textbf{12}     & \textbf{0.556} & \textbf{0.063} & \textbf{0.938}     \\
			\midrule 
			\multirow{7}{*}{10} & PC-Avg          & 2.9     & 1.3   & 12.4   & 16.6   & 0.433 & 0.260 & 0.740     \\
			& PC-Best         & 2       & \textbf{0}     & 11     & 13     & 0.519 & 0.125 & 0.875     \\
			& FedPC-Simple-I  & 2       & 1     & 10     & 13     & 0.556 & 0.167 & 0.833     \\
			& FedPC-Simple-II & 2       & 16    & \textbf{8}      & 26     & \textbf{0.630} & 0.514 & 0.486     \\
			& NOTEARS-Avg     & \textbf{0}       & \textbf{0}     & 25     & 25     & 0.074 & \textbf{0.000} & \textbf{1.000}     \\
			& NOTEARS-ADMM    & 3       & 33    & 19     & 55     & 0.185 & 0.878 & 0.122     \\
			& {FedDAG} & {26}                             & {28}                           & {0}                           & {54}                         & {1.000}                      & {0.667}                     & {0.333}                           \\
			
			& FedPC           & \textbf{0}       & \textbf{0}     & 11     & \textbf{11}     & 0.593 & \textbf{0.000} & \textbf{1.000}     \\
			\midrule 
			\multirow{7}{*}{15} & PC-Avg          & 2       & 1.667 & 13.667 & 17.333 & 0.420 & 0.235 & 0.765     \\
			& PC-Best         & 1       & 1     & 12     & 14     & 0.519 & 0.125 & 0.875     \\
			& FedPC-Simple-I  & 1       & 1     & 11     & 13     & 0.556 & 0.118 & 0.882     \\
			& FedPC-Simple-II & 2       & 16    & \textbf{9}      & 27     & \textbf{0.593} & 0.529 & 0.471     \\
			& NOTEARS-Avg     & \textbf{0}       & \textbf{0}     & 25     & 25     & 0.074 & \textbf{0.000} & \textbf{1.000}     \\
			& NOTEARS-ADMM    & 3       & 29    & 21     & 53     & 0.111 & 0.914 & 0.086     \\
			& {FedDAG} & {26}                             & {28}                           & {0}                           & {54}                         & {1.000}                      & {0.667}                     & {0.333}                           \\
			& FedPC           & 1       & \textbf{0}     & 10     & \textbf{11}     & \textbf{0.593} & 0.059 & 0.941     \\
			\bottomrule 
	\end{tabular}}
\end{table*}

\begin{table*}[htbp]
	\caption{Details of continuous datasets: 50 nodes}
	\label{tbl-50n}
	\centering
	\scriptsize
	\setlength{\tabcolsep}{2mm}{
		\begin{tabular}{ccccccccc}
			\toprule 
			Number of clients   & Algorithm    & Reverse($\downarrow$) & Extra($\downarrow$)  & Miss($\downarrow$)  & SHD($\downarrow$)    & TPR ($\uparrow$)   & FDR($\downarrow$)   & Precision($\uparrow$)\\
			\midrule 
			\multirow{7}{*}{3}  & PC-Avg          & 6       & 18.667 & 28     & 52.667 & 0.521 & 0.399 & 0.601     \\
			& PC-Best         & 6       & 13     & 24     & 43     & 0.577 & 0.317 & 0.683     \\
			& FedPC-Simple-I  & 6       & 50     & \textbf{16}     & 72     & 0.690 & 0.533 & 0.467     \\
			& FedPC-Simple-II & 7       & 76     & 13     & 96     & \textbf{0.718} & 0.619 & 0.381     \\
			& NOTEARS-Avg     & 9       & 6      & 50     & 65     & 0.169 & 0.556 & 0.444     \\
			& NOTEARS-ADMM    & 14      & 29     & 32     & 75     & 0.352 & 0.632 & 0.368     \\
			& {FedDAG} & {71}                             & {72}                           & {0}                           & {143}                        & {1.000}                      & {0.668}                     & {0.332}                           \\
			& FedPC           & \textbf{2}       & \textbf{3}      & 32     & \textbf{37}     & 0.521 & \textbf{0.119} & \textbf{0.881}     \\
			\midrule 
			\multirow{7}{*}{5}  & PC-Avg          & 3.8     & 12     & 28.4   & 44.2   & 0.546 & 0.283 & 0.717     \\
			& PC-Best         & 3       & 4      & 25     & 32     & 0.606 & 0.140 & 0.860     \\
			& FedPC-Simple-I  & 2       & 7      & 19     & \textbf{28}     & \textbf{0.704} & 0.153 & 0.847     \\
			& FedPC-Simple-II & 6       & 33     & \textbf{15}     & 54     & \textbf{0.704} & 0.438 & 0.562     \\
			& NOTEARS-Avg     & 4       & \textbf{1}      & 56     & 61     & 0.155 & 0.313 & 0.688     \\
			& NOTEARS-ADMM    & 12      & 43     & 36     & 91     & 0.324 & 0.705 & 0.295     \\
			& {FedDAG} & {71}                             & {72}                           & {0}                           & {143}                        & {1.000}                      & {0.668}                     & {0.332}                           \\
			& FedPC           & \textbf{0}       & 3      & 25     & \textbf{28}     & 0.648 & \textbf{0.061} & \textbf{0.939}     \\
			\midrule 
			\multirow{7}{*}{10} & PC-Avg          & 3.6     & 6.5    & 31     & 41.1   & 0.513 & 0.216 & 0.784     \\
			& PC-Best         & 1       & 3      & 28     & 32     & 0.592 & 0.087 & 0.913     \\
			& FedPC-Simple-I  & 1       & 6      & 25     & 32     & 0.634 & 0.135 & 0.865     \\
			& FedPC-Simple-II & 6       & 44     & \textbf{13}     & 63     & \textbf{0.732} & 0.490 & 0.510     \\
			& NOTEARS-Avg     & 5       & \textbf{0}      & 56     & 61     & 0.141 & 0.333 & 0.667     \\
			& NOTEARS-ADMM    & 9       & 49     & 44     & 102    & 0.254 & 0.763 & 0.237     \\
			& {FedDAG} & {71}                             & {72}                           & {0}                           & {143}                        & {1.000}                      & {0.668}                     & {0.332}                           \\
			& FedPC           & \textbf{0}       & 4      & 22     & \textbf{26}     & 0.690 & \textbf{0.075} & \textbf{0.925}     \\
			\midrule 
			\multirow{7}{*}{15} & PC-Avg          & 3.333   & 5.733  & 33.733 & 42.8   & 0.478 & 0.209 & 0.791     \\
			& PC-Best         & 4       & 3      & 30     & 37     & 0.521 & 0.159 & 0.841     \\
			& FedPC-Simple-I  & 1       & 5      & 30     & 36     & 0.563 & 0.130 & 0.870     \\
			& FedPC-Simple-II & 4       & 51     & \textbf{13}     & 68     & \textbf{0.761} & 0.505 & 0.495     \\
			& NOTEARS-Avg     & 5       & \textbf{1}      & 58     & 64     & 0.113 & 0.429 & 0.571     \\
			& NOTEARS-ADMM    & 8       & 40     & 46     & 94     & 0.239 & 0.738 & 0.262     \\
			& {FedDAG} & {71}                             & {72}                           & {0}                           & {143}                        & {1.000}                      & {0.668}                     & {0.332}                           \\
			& FedPC           & \textbf{0}       & 7      & 21     & \textbf{28}     & 0.704 & \textbf{0.123} & \textbf{0.877}     \\
			\bottomrule 
	\end{tabular}}
\end{table*}

\begin{table*}[htbp]
	\caption{Details of continuous datasets: 100 nodes}
	\label{tbl-100n}
	\centering
	\scriptsize
	\setlength{\tabcolsep}{2mm}{
		\begin{tabular}{ccccccccc}
			\toprule 
			Number of clients   & Algorithm    & Reverse($\downarrow$) & Extra($\downarrow$)  & Miss($\downarrow$)  & SHD($\downarrow$)    & TPR ($\uparrow$)   & FDR($\downarrow$)   & Precision($\uparrow$)\\
			\midrule 
			\multirow{7}{*}{3}  & PC-Avg          & 7.667   & 38.667 & 63.333 & 109.667 & 0.551 & 0.339 & 0.661     \\
			& PC-Best         & 6       & 36     & 53     & 95      & 0.627 & 0.298 & 0.702     \\
			& FedPC-Simple-I  & 6       & 94     & 48     & 148     & 0.658 & 0.490 & 0.510     \\
			& FedPC-Simple-II & 7       & 218    & \textbf{46}     & 271     & \textbf{0.665} & 0.682 & 0.318     \\
			& NOTEARS-Avg     & 14      & 16     & 99     & 129     & 0.285 & 0.400 & 0.600     \\
			& NOTEARS-ADMM    & 19      & 26     & 102    & 147     & 0.234 & 0.549 & 0.451     \\
			& {FedDAG} & {-}                              & {-}                            & {-}                           & {-}                          & {-}                           & {-}                          & {-}                                \\
			& FedPC           & \textbf{5}       & \textbf{9}      & 80     & \textbf{94}      & 0.462 & \textbf{0.161} & \textbf{0.839}     \\
			\midrule 
			\multirow{7}{*}{5}  & PC-Avg          & 7.6     & 28     & 68.6   & 104.2   & 0.518 & 0.298 & 0.702     \\
			& PC-Best         & 10      & 19     & 57     & 86      & 0.576 & 0.242 & 0.758     \\
			& FedPC-Simple-I  & \textbf{4}       & 17     & 57     & 78      & 0.614 & 0.178 & 0.822     \\
			& FedPC-Simple-II & 10      & 70     & \textbf{47}     & 127     & \textbf{0.639} & 0.442 & 0.558     \\
			& NOTEARS-Avg     & 13      & 11     & 100    & 124     & 0.291 & 0.343 & 0.657     \\
			& NOTEARS-ADMM    & 20      & 19     & 109    & 148     & 0.184 & 0.574 & 0.426     \\
			& {FedDAG} & {-}                              & {-}                            & {-}                           & {-}                          & {-}                           & {-}                          & {-}                                \\
			& FedPC           & 7       & \textbf{8}      & 61     & \textbf{76}      & 0.570 & \textbf{0.143} & \textbf{0.857}     \\
			\midrule 
			\multirow{7}{*}{10} & PC-Avg          & \textbf{6.8}     & 18.3   & 70.8   & 95.9    & 0.509 & 0.237 & 0.763     \\
			& PC-Best         & 7       & 11     & 66     & 84      & 0.538 & \textbf{0.175} & \textbf{0.825}     \\
			& FedPC-Simple-I  & 7       & 16     & 54     & 77      & 0.614 & 0.192 & 0.808     \\
			& FedPC-Simple-II & 7       & 113    & \textbf{44}     & 164     & \textbf{0.677} & 0.529 & 0.471     \\
			& NOTEARS-Avg     & 13      & \textbf{9}      & 102    & 124     & 0.272 & 0.338 & 0.662     \\
			& NOTEARS-ADMM    & 19      & 12     & 115    & 146     & 0.152 & 0.564 & 0.436     \\
			& {FedDAG} & {-}                              & {-}                            & {-}                           & {-}                          & {-}                           & {-}                          & {-}                                \\
			& FedPC           & 10      & 11     & 54     & \textbf{75}      & 0.595 & 0.183 & 0.817     \\
			\midrule 
			\multirow{7}{*}{15} & PC-Avg          & 6.533   & 16.6   & 78.6   & 101.733 & 0.461 & 0.239 & 0.761     \\
			& PC-Best         & 6       & \textbf{9}      & 75     & 90      & 0.487 & \textbf{0.163} & \textbf{0.837}     \\
			& FedPC-Simple-I  & \textbf{5}       & 13     & 71     & 89      & 0.519 & 0.180 & 0.820     \\
			& FedPC-Simple-II & 13      & 100    & \textbf{46}     & 159     & \textbf{0.627} & 0.533 & 0.467     \\
			& NOTEARS-Avg     & 12      & 10     & 107    & 129     & 0.253 & 0.355 & 0.645     \\
			& NOTEARS-ADMM    & 16      & \textbf{9}      & 122    & 147     & 0.127 & 0.556 & 0.444     \\
			& {FedDAG} & {-}                              & {-}                            & {-}                           & {-}                          & {-}                           & {-}                          & {-}                                \\
			& FedPC           & 9       & 15     & 57     & \textbf{81}      & 0.582 & 0.207 & 0.793     \\
			\bottomrule 
	\end{tabular}}
\end{table*}

\begin{table*}[htbp]
	\caption{Details of continuous datasets: 300 nodes}
	\label{tbl-300n}
	\centering
	\scriptsize
	\setlength{\tabcolsep}{2mm}{
		\begin{tabular}{ccccccccc}
			\toprule 
			Number of clients   & Algorithm    & Reverse($\downarrow$) & Extra($\downarrow$)  & Miss($\downarrow$)  & SHD($\downarrow$)    & TPR ($\uparrow$)   & FDR($\downarrow$)   & Precision($\uparrow$)\\
			\midrule 
			\multirow{7}{*}{3}  & PC-Avg          & 14      & 139.333 & 159     & 312.333 & 0.586 & 0.343 & 0.657     \\
			& PC-Best         & 14      & 47      & 168     & 229     & 0.565 & 0.205 & 0.795     \\
			& FedPC-Simple-I  & 6       & 393     & 116     & 515     & 0.708 & 0.574 & 0.426     \\
			& FedPC-Simple-II & \textbf{0}       & 2427    & \textbf{88}      & 2515    & \textbf{0.789} & 0.880 & 0.120     \\
			& NOTEARS-Avg     & 23      & \textbf{5}       & 311     & 339     & 0.203 & 0.248 & 0.752     \\
			& NOTEARS-ADMM    & 74      & 64      & 226     & 364     & 0.282 & 0.539 & 0.461     \\
			& {FedDAG} & {-}                              & {-}                            & {-}                           & {-}                          & {-}                           & {-}                          & {-}                                \\
			& FedPC           & 4       & 9       & 187     & \textbf{200}     & 0.543 & \textbf{0.054} & \textbf{0.946}     \\
			\midrule 
			\multirow{7}{*}{5}  & PC-Avg          & 12.2    & 109.4   & 160.4   & 282     & 0.587 & 0.316 & 0.684     \\
			& PC-Best         & 11      & 90      & 130     & 231     & 0.663 & 0.267 & 0.733     \\
			& FedPC-Simple-I  & \textbf{5}       & 44      & 128     & 177     & \textbf{0.682} & 0.147 & 0.853     \\
			& FedPC-Simple-II & 28      & 353     & \textbf{109}     & 490     & 0.672 & 0.576 & 0.424     \\
			& NOTEARS-Avg     & 23      & \textbf{10}      & 312     & 345     & 0.199 & 0.284 & 0.716     \\
			& NOTEARS-ADMM    & 50      & 40      & 253     & 343     & 0.275 & 0.439 & 0.561     \\
			& {FedDAG} & {-}                              & {-}                            & {-}                           & {-}                          & {-}                           & {-}                          & {-}                                \\
			& FedPC           & 6       & 26      & 144     & \textbf{176}     & 0.641 & \textbf{0.107} & \textbf{0.893}     \\
			\midrule 
			\multirow{7}{*}{10} & PC-Avg          & 13.5    & 67.4    & 175.4   & 256.3   & 0.548 & 0.260 & 0.740     \\
			& PC-Best         & 11      & 51      & 162     & 224     & 0.586 & 0.202 & 0.798     \\
			& FedPC-Simple-I  & \textbf{3}       & 30      & 139     & \textbf{172}     & 0.660 & \textbf{0.107} & \textbf{0.893}     \\
			& FedPC-Simple-II & 18      & 756     & \textbf{97}      & 871     & \textbf{0.725} & 0.719 & 0.281     \\
			& NOTEARS-Avg     & 23      & \textbf{6}       & 321     & 350     & 0.179 & 0.279 & 0.721     \\
			& NOTEARS-ADMM    & 45      & 25      & 283     & 353     & 0.215 & 0.438 & 0.563     \\
			& {FedDAG} & {-}                              & {-}                            & {-}                           & {-}                          & {-}                           & {-}                          & {-}                                \\
			& FedPC           & 4       & 31      & 140     & 175     & 0.656 & 0.113 & 0.887     \\
			\midrule 
			\multirow{7}{*}{15} & PC-Avg          & 15.667  & 56.733  & 185.333 & 257.733 & 0.519 & 0.250 & 0.750     \\
			& PC-Best         & 12      & 49      & 166     & 227     & 0.574 & 0.203 & 0.797     \\
			& FedPC-Simple-I  & \textbf{8}       & 21      & 162     & 191     & 0.593 & \textbf{0.105} & \textbf{0.895}     \\
			& FedPC-Simple-II & 32      & 375     & \textbf{110}     & 517     & \textbf{0.660} & 0.596 & 0.404     \\
			& NOTEARS-Avg     & 24      & \textbf{6}       & 324     & 354     & 0.167 & 0.300 & 0.700     \\
			& NOTEARS-ADMM    & 42      & 21      & 291     & 354     & 0.203 & 0.426 & 0.574     \\
			& {FedDAG} & {-}                              & {-}                            & {-}                           & {-}                          & {-}                           & {-}                          & {-}                                \\
			& FedPC           & 10      & 40      & 133     & \textbf{183}     & 0.658 & 0.154 & 0.846     \\
			\bottomrule 
	\end{tabular}}
\end{table*}

\subsection*{S-1-4: Experiment results on real data.}
To compare the performance of our proposed framework on a real dataset, we consider a real bioinformatics dataset Sachs. Sachs is a protein signaling network expressing the level of different proteins and phospholipids in human cells.
It is commonly viewed as a benchmark graphical model with 11 nodes and 17 edges. In our experiments, we adopt the observational data with 853 samples.

Among all methods in the experiments, as shown in Table \ref{tbl-sachs}, the performance of FedPC-Simple-I, FedPC-Simple-II, PC-Avg and PC-Best are competitive, but FedPC achieves the best performance with highest precision values and lowest FDR vlaues. It means that our method gets a more accurate DAG and identifies more correct directed edges than its rivals.
Continuous optimization approaches, i.e. NOTEARS-Avg and NOTEARS-ADMM, achieve a high TPR, it also learns many extra edges. In addition, we also observe that the performance of continuous optimization approaches is comparable to that of other comparison methods on real data, whereas generally worse than that of traditional methods on benchmark data.

\begin{table*}[htbp]
	\caption{Details of the real dataset}
	\label{tbl-sachs}
	\centering
	\scriptsize
	\setlength{\tabcolsep}{2mm}{
		\begin{tabular}{ccccccccc}
			\toprule 
			Number of clients   & Algorithm    & Reverse($\downarrow$) & Extra($\downarrow$)  & Miss($\downarrow$)  & SHD($\downarrow$)    & TPR ($\uparrow$)   & FDR($\downarrow$)   & Precision($\uparrow$)\\
			\midrule 
			\multirow{7}{*}{3}  & PC-Avg          & 4.333   & \textbf{0}     & 9.667  & 14   & 0.176 & 0.595 & 0.405     \\
			& PC-Best         & 4       & \textbf{0}     & 9      & \textbf{13}   & 0.235 & \textbf{0.500} & \textbf{0.500}     \\
			& FedPC-Simple-I  & 5       & \textbf{0}     & 9      & 14   & 0.176 & 0.625 & 0.375     \\
			& FedPC-Simple-II & \textbf{3}       & 8     & 8      & 19   & \textbf{0.353} & 0.647 & 0.353     \\
			& NOTEARS-Avg     & 7       & 5     & \textbf{4}      & 16   & \textbf{0.353} & 0.667 & 0.333     \\
			& NOTEARS-ADMM    & 7       & 10    & \textbf{4}      & 21   & \textbf{0.353} & 0.739 & 0.261     \\
			& {FedDAG} & {6}                              & {10}                           & {5}                           & {21}                         & {0.353}                    & {0.727}                   & {0.273}                         \\
			& FedPC           & 4       & \textbf{0}     & 9      & \textbf{13}   & 0.235 & \textbf{0.500} & \textbf{0.500}     \\
			\midrule 
			\multirow{7}{*}{5}  & PC-Avg          & 3.8     & 0.2   & 10.4   & 14.4 & 0.165 & 0.556 & 0.444     \\
			& PC-Best         & 4       & \textbf{0}     & 10     & 14   & 0.176 & 0.571 & 0.429     \\
			& FedPC-Simple-I  & 5       & \textbf{0}     & 9      & 14   & 0.176 & 0.625 & 0.375     \\
			& FedPC-Simple-II & 5       & 4     & 9      & 18   & 0.176 & 0.750 & 0.250     \\
			& NOTEARS-Avg     & 7       & 6     & 5      & 18   & 0.294 & 0.722 & 0.278     \\
			& NOTEARS-ADMM    & 8       & 10    & \textbf{2}      & 20   & \textbf{0.412} & 0.720 & 0.280     \\
			& {FedDAG} & {7}                              & {10}                           & {4}                           & {21}                         & {0.353}                       & {0.739}                      & {0.261}                            \\
			& FedPC           & \textbf{2}       & \textbf{0}     & 9      & \textbf{11}   & 0.353 & \textbf{0.250} & \textbf{0.750}     \\
			\midrule 
			\multirow{7}{*}{10} & PC-Avg          & 2       & 0     & 11.7   & 13.7 & 0.194 & 0.355 & 0.645     \\
			& PC-Best         & \textbf{0}       & \textbf{0}     & 12     & 12   & 0.294 & \textbf{0.000} & \textbf{1.000}     \\
			& FedPC-Simple-I  & 3       & \textbf{0}     & 10     & 13   & 0.235 & 0.429 & 0.571     \\
			& FedPC-Simple-II & 3       & 10    & 8      & 21   & 0.353 & 0.684 & 0.316     \\
			& NOTEARS-Avg     & 7       & 9     & 4      & 20   & 0.353 & 0.727 & 0.273     \\
			& NOTEARS-ADMM    & 8       & 13    & \textbf{3}      & 24   & 0.353 & 0.778 & 0.222     \\
			& {FedDAG} & {7}                              & {12}                           & {2}                           & {21}                         & {0.471}                       & {0.704}                      & {0.296}                            \\
			& FedPC           & \textbf{0}       & \textbf{0}     & 10     & \textbf{10}   & \textbf{0.412} & \textbf{0.000} & \textbf{1.000}     \\
			\midrule 
			\multirow{7}{*}{15} & PC-Avg          & 1.533   & 0.133 & 12.133 & 13.8 & 0.196 & 0.313 & 0.687     \\
			& PC-Best         & \textbf{0}       & 1     & 12     & 13   & 0.294 & \textbf{0.167} & \textbf{0.833}     \\
			& FedPC-Simple-I  & 2       & \textbf{0}     & 12     & 14   & 0.176 & 0.400 & 0.600     \\
			& FedPC-Simple-II & 5       & 6     & 8      & 19   & 0.235 & 0.733 & 0.267     \\
			& NOTEARS-Avg     & 7       & 12    & \textbf{3}      & 22   & 0.412 & 0.731 & 0.269     \\
			& NOTEARS-ADMM    & 6       & 15    & \textbf{3}      & 24   & \textbf{0.471} & 0.724 & 0.276     \\
			& {FedDAG} & {6}                              & {14}                           & {4}                           & {24}                         & {0.412}                       & {0.741}                      & {0.259}                            \\
			& FedPC           & 2       & \textbf{0}     & 10     & \textbf{12}   & 0.294 & 0.286 & 0.714     \\
			\bottomrule 
	\end{tabular}}
\end{table*}

\begin{table*}[htbp]
	\caption{Details of the results on nonlinear datasets}
	\label{tbl-nlnr}
	\centering
	\scriptsize
	\setlength{\tabcolsep}{2mm}{
		\begin{tabular}{ccccccc}
			\toprule 
			Nonlinear datasets         & Number of clients   & Algorithms                   & SHD($\downarrow$)      & TPR ($\uparrow$)         & Precision($\uparrow$)    & F1($\uparrow$)           \\
			\midrule 
			\multirow{12}{*}{10 nodes} & \multirow{3}{*}{3}  & FedDAG                       & 5           & 0.818          & 0.692          & 0.750          \\ 
			&                     & NOTEARS-MLP-ADMM             & 27          & 0.091          & 0.040          & 0.056          \\ 
			&                     & FedPC-KBT & \textbf{4}  & \textbf{0.909} & \textbf{0.714} & \textbf{0.800} \\ \cmidrule{3-7}
			& \multirow{3}{*}{5}  & FedDAG                       & 4           & 0.636          & \textbf{0.875} & 0.737          \\
			&                     & NOTEARS-MLP-ADMM             & 15          & 0.182          & 0.182          & 0.182          \\
			&                     & FedPC-KBT & \textbf{2}  & \textbf{0.909} & 0.833          & \textbf{0.870} \\ \cmidrule{3-7}
			& \multirow{3}{*}{10} & FedDAG                       & 7           & 0.364          & 0.667          & 0.471          \\
			&                     & NOTEARS-MLP-ADMM             & 14          & 0.182          & 0.222          & 0.200          \\
			&                     & FedPC-KBT & \textbf{2}  & \textbf{0.818} & \textbf{0.818} & \textbf{0.818} \\ \cmidrule{3-7}
			& \multirow{3}{*}{15} & FedDAG                       & 7           & 0.364          & 0.667          & 0.471          \\
			&                     & NOTEARS-MLP-ADMM             & 14          & 0.091          & 0.111          & 0.100          \\
			&                     & FedPC-KBT & \textbf{3}  & \textbf{0.727} & \textbf{0.889} & \textbf{0.800} \\ 
			\midrule 
			\multirow{12}{*}{20 nodes} & \multirow{3}{*}{3}  & FedDAG                       & 24          & 0.143          & 0.500          & 0.222          \\
			&                     & NOTEARS-MLP-ADMM             & 30          & 0.286          & 0.348          & 0.314          \\
			&                     & FedPC-KBT &  \textbf{11}           &  \textbf{0.786}               &  \textbf{0.667}               &   \textbf{0.721}              \\ \cmidrule{3-7}
			& \multirow{3}{*}{5}  & FedDAG                       & 26          & 0.071          & 0.333          & 0.118          \\
			&                     & NOTEARS-MLP-ADMM             & 36          & 0.250          & 0.250          & 0.250          \\
			&                     & FedPC-KBT & \textbf{4}  & \textbf{0.893} & \textbf{0.862} & \textbf{0.877} \\ \cmidrule{3-7}
			& \multirow{3}{*}{10} & FedDAG                       & 27          & 0.036          & 1.000          & 0.069         \\
			&                     & NOTEARS-MLP-ADMM             & 31          & 0.071          & 0.250          & 0.111          \\
			&                     & FedPC-KBT & \textbf{7}  & \textbf{0.786} & \textbf{0.815} & \textbf{0.800} \\ \cmidrule{3-7}
			& \multirow{3}{*}{15} & FedDAG                       & 27          & 0.036          & 0.500          & 0.067          \\
			&                     & NOTEARS-MLP-ADMM             & 30          & 0.071          & 0.222          & 0.108          \\
			&                     & FedPC-KBT & \textbf{13} & \textbf{0.571} & \textbf{0.696} & \textbf{0.628} \\ 
			\bottomrule 
	\end{tabular}}
\end{table*}

\section*{S-2: An illustrative example of the privacy protection strategy employed by FedPC}\label{secS-2} 
\begin{figure*}[t]
	\centering
	\includegraphics[width=0.95\textwidth]{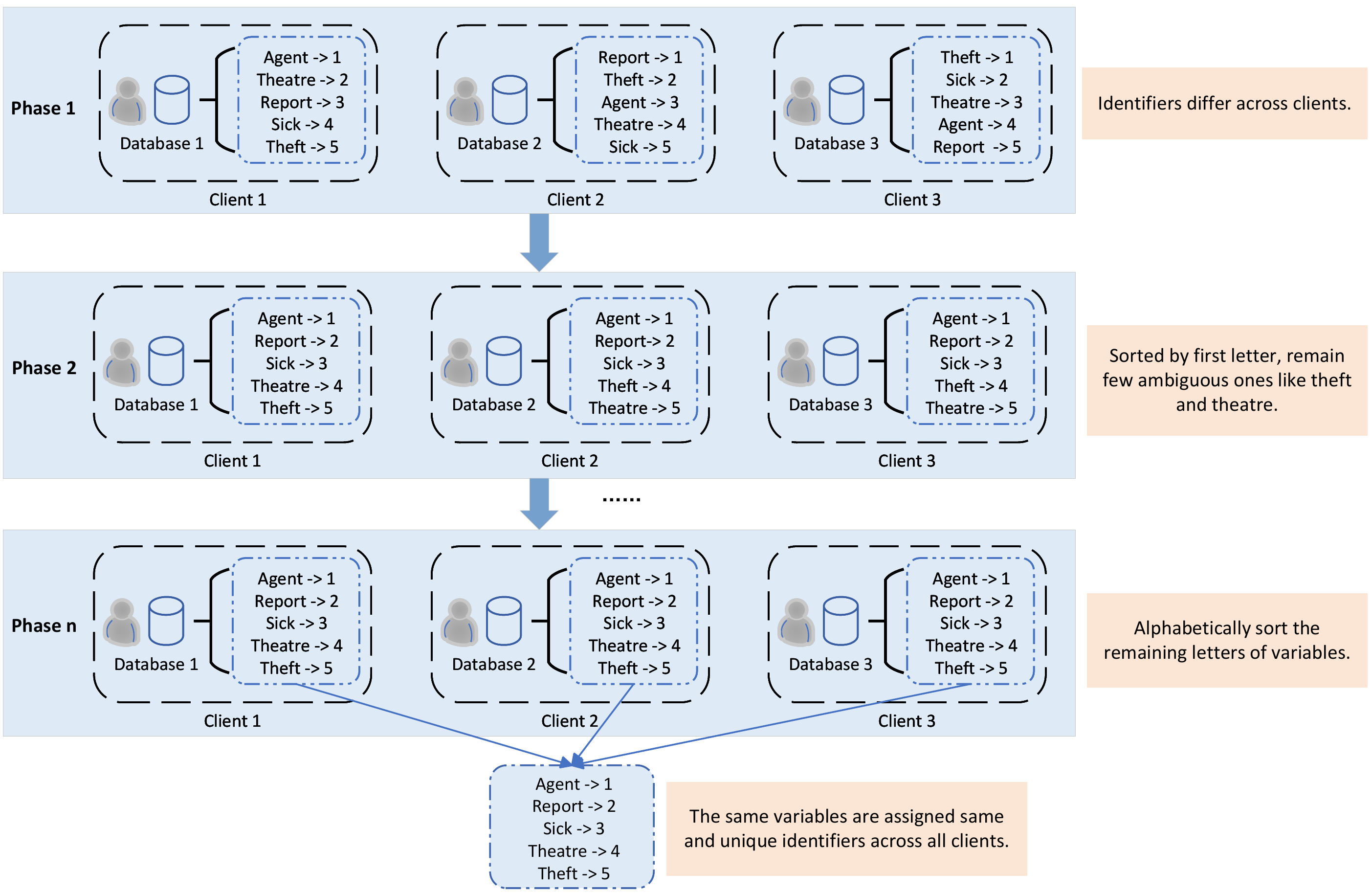}\\
	\caption{An easily implementable privacy protection strategy in the FedPC algorithm.}
	\label{pri-prot-strategy}
\end{figure*}

As shown in Fig.~\ref{pri-prot-strategy}, we firstly sort the first letters of all variables on each client in alphabetical order to establish a correspondence between variables and identifiers. However, this correspondence may encounter instances of ambiguity, exemplified by the case of "Theatre" and "Theft". Because both words share identical first and second letters, sorting based solely on the initial letter fails to determine a unique and unambiguous mapping between variables and identifiers. Then we sort the remaining letters of each variable until all letters have been sorted. The same feature space among clients ensures that the resulting correspondence between variables and identifiers remains consistent. Furthermore, this method effectively safeguards against the server's access to the semantic information associated with each variable, thereby mitigating potential risks of privacy leakage.

\end{appendices}

\end{document}